\documentclass[letterpaper, 10 pt, conference]{ieeeconf}  

\IEEEoverridecommandlockouts                              
\usepackage[utf8]{inputenc} 
\usepackage[T1]{fontenc}    

\usepackage{etoolbox}

\newtoggle{isArxiv}

\toggletrue{isArxiv} 

\newcommand{\arxiv}[1]{%
    \iftoggle{isArxiv}{#1}{}%
}

\newcommand{\icra}[1]{%
    \iftoggle{isArxiv}{}{#1}%
}

\iftoggle{isArxiv}{
  
}{
}

\makeatletter
\let\NAT@parse\undefined
\makeatother

\usepackage[hidelinks]{hyperref}   

\iftoggle{isArxiv}{%
  \usepackage[backend=biber, style=numeric-comp, sorting=ynt, natbib=true,
              eprint=true, doi=true, url=false, isbn=false,
              maxbibnames=99]{biblatex}%
}{%
    \usepackage[
      backend=biber,
      style=ieee,
      citestyle=ieee-comp,
      sorting=none,
      natbib=true,
      doi=false,
      url=false,
      eprint=true,      
      isbn=false
    ]{biblatex}
    \DeclareFieldFormat{eprint:arxiv}{arXiv preprint arXiv:#1}
}
\icra{\setlength{\bibitemsep}{0pt plus .3ex}}
\usepackage{url}            
\usepackage{booktabs}       
\usepackage{amsfonts}       
\usepackage{nicefrac}       
\usepackage{microtype}      
\usepackage{xcolor}         

\usepackage{graphicx}
\usepackage{subcaption}
\usepackage{algorithm}
\usepackage[noEnd,commentColor=black]{algpseudocodex}
\usepackage[most]{tcolorbox}
\usepackage{float}
\usepackage{wrapfig}
\usepackage{lipsum}
\usepackage{colortbl}
\usepackage{soul}
\let\labelindent\relax      
\usepackage{enumitem}
\usepackage{multirow}
\usepackage{makecell}
\usepackage{tablefootnote}
\usepackage[para,online,flushleft]{threeparttable}
\usepackage{siunitx}
\usepackage[normalem]{ulem}   
\usepackage{amsmath}
\usepackage{xspace}
\usepackage[nameinlink,noabbrev]{cleveref}   
\usepackage{cuted}
\usepackage{caption}
\usepackage{amsfonts} 

\title{\LARGE \bf
Reinforcement Learning for Real-Time Vision-Language-Action Policies
}

\newcommand{\symfootnotetext}[2]{%
  \begingroup
  \renewcommand{\thefootnote}{#1}%
  \footnotetext{#2}%
  \endgroup
}

\arxiv{
\author{
 Perry Dong$^{*, \dagger}$ \quad Kuo-Han Hung$^{*}$ \quad Dorsa Sadigh \quad Chelsea Finn \\[8pt]
  Stanford University \\[2pt]
  \href{https://pd-perry.github.io/real-time-expo-ft/}{\texttt{https://pd-perry.github.io/real-time-expo-ft/}}
}
}

\icra{
\author{
 Anonymous Authors
}
}

\newcommand{\methodname}{Real-Time EXPO-FT\xspace}

\begin{document}
\maketitle
\symfootnotetext{*}{Equal contributions. } 
\symfootnotetext{$\dagger$}{Corresponding author: \texttt{perryd@stanford.edu}}

\begin{strip}
    \icra{\vspace{-12mm}}
    \arxiv{\vspace{-17mm}}  
    \centering
    \arxiv{\includegraphics[width=\textwidth]{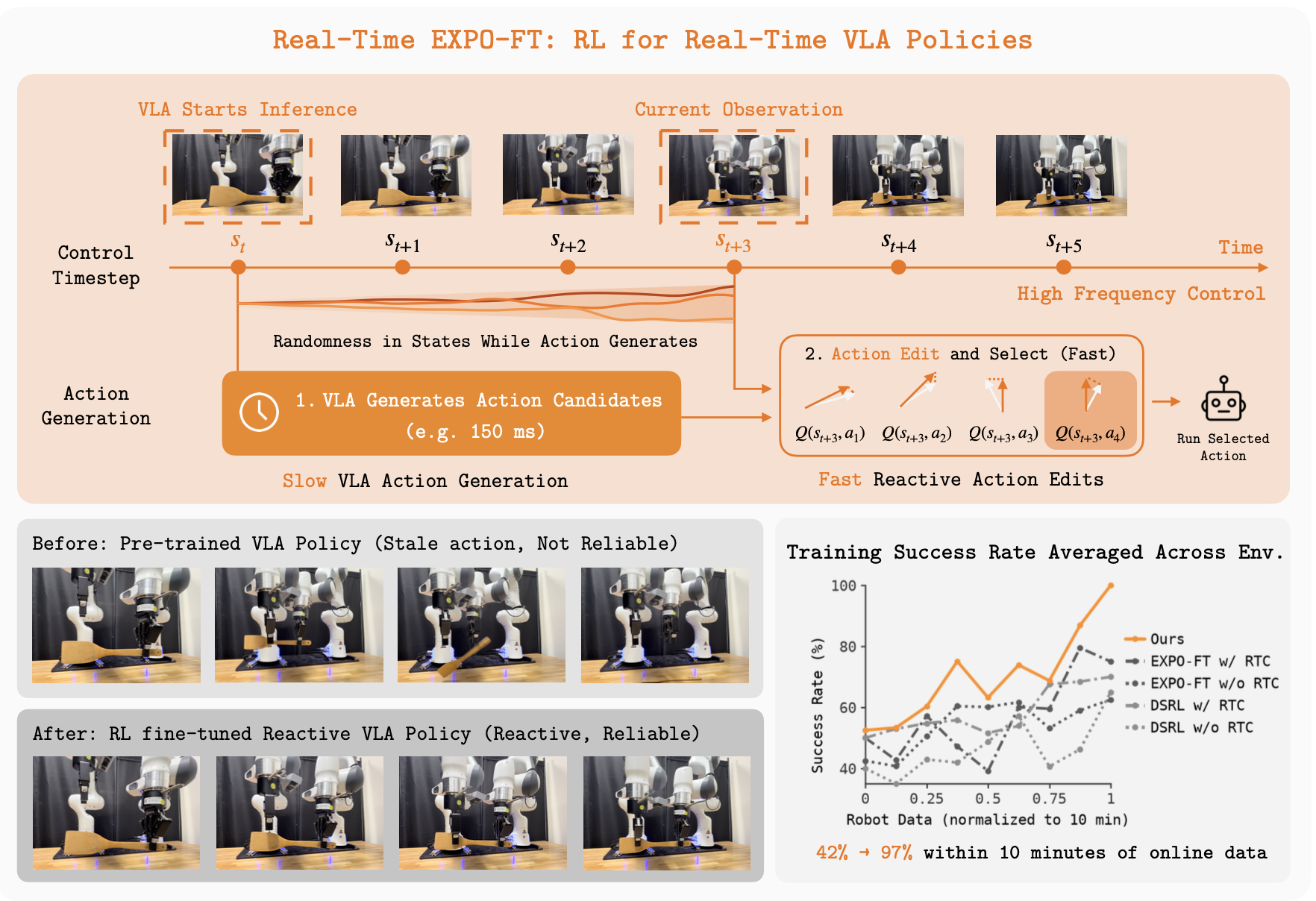}}
    \icra{\includegraphics[width=\textwidth]{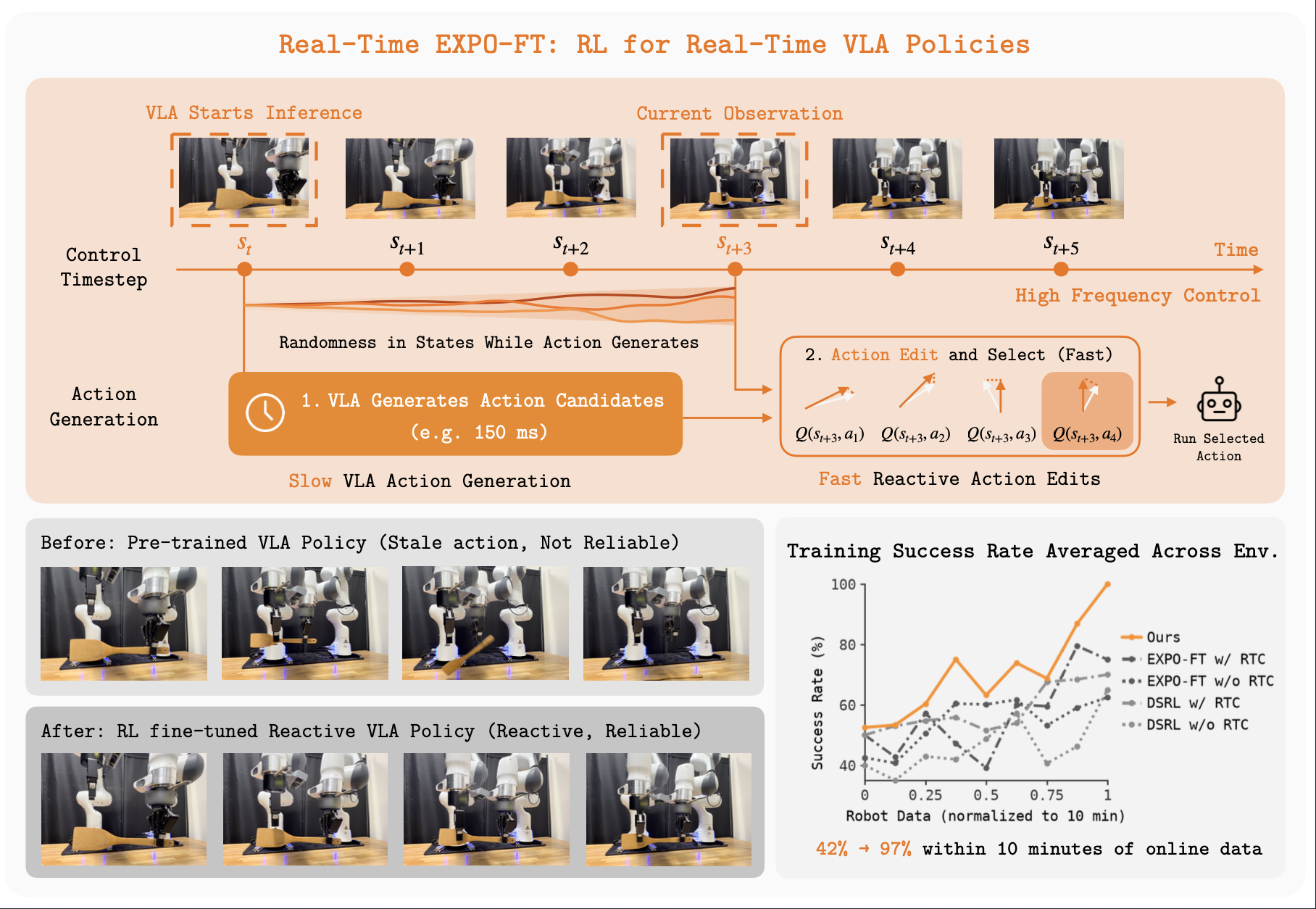}}
    \captionof{figure}{\textbf{Overview of \methodname.} \methodname addresses the latency of large VLA models by combining slow action generation with fast reactive control. \textbf{Bottom left:} Task success rates before and after applying \methodname. \textbf{Bottom right:} Training success rates on real-world tasks compared with baselines.}
    \label{fig:main}
\end{strip}

\thispagestyle{empty}
\pagestyle{empty}

\newtcolorbox{draftbox}{
    colback=yellow!10,
    colframe=orange
}

\begin{abstract}
Reinforcement learning fine-tuning on top of large, pretrained Vision-Language-Action (VLA) models offers promise for highly reliable robot deployment. However, because of their scale, modern VLA models suffer from high inference latency, so the observation used to select an action is often stale by execution time, creating a distribution shift that can substantially degrade reliability and performance. Prior work has explored asynchronous policy execution to reduce the effect of latency, but these methods are mostly built on imitation learning and offer no mechanism for moving beyond the training distribution toward higher reliability. We close this gap by enabling RL fine-tuning that meets the real-time control requirements of dynamic real-world manipulation. Our approach builds on EXPO-FT, a framework for sample-efficient, reliable VLA fine-tuning with reinforcement learning, and decouples slow, expressive action generation from fast, reactive action edits: a large pretrained VLA proposes action chunks using its strong behavior prior, while a lightweight edit policy performs fast, reactive decision-making by editing actions in response to changes in state, conditioned on the latest observation. We instantiate this as \methodname, an RL framework for finetuning real-time VLA policies. On the Kinetix benchmark, \methodname enables a delayed policy to achieve the best performance among delayed and non-delayed methods in 10 out of 10 environments. On four dynamic real-world tasks—robot object passing, ball balancing, table soccer kicking, and dynamic object picking—with online robot data capped at 10 minutes, \methodname improves average policy performance from 42\% to 97\%, all without human intervention, demonstrating rapid, sample-efficient adaptation to challenging real-world dynamics. \\ 
\icra{
Website: \href{https://real-time-rl.pages.dev/}{\textcolor{orange}{https://real-time-rl.pages.dev/}}
}

\end{abstract}

\section{Introduction}

Reinforcement learning fine-tuning on top of large, pretrained vision-language-action (VLA) models offers promise toward highly reliable robot deployment in the real world~\citep{dong2026expoftsampleefficientreinforcementlearning,intelligence2025pi06vlalearnsexperience}. However, the scale behind pretrained VLA models that makes it a strong behavior prior cuts both ways. While the large capacity enables complex, multi-step behaviors across diverse tasks and embodiments, it also inflates inference latency, and the physical world does not pause while the robot computes its next action. As a result, for the most capable models, the observation used to infer an action is often not the observation at the time of execution, creating a distribution shift that can substantially degrade reliability and performance. In this work, we study RL fine-tuning of VLA policies for real-time execution, trained explicitly considering inference latency of a large VLA model so that the policy can retain maximal reactivity to changes in the environment while achieving higher reliability with reinforcement learning. 

Prior work on asynchronous policy execution has explored alternative inference schemes~\citep{black2025realtime, tang2026vlashrealtimevlasfuturestateaware}, alternative training schemes~\citep{black2025trainingtimeactionconditioningefficient}, or both~\citep{sendai2025leaveobservationbehindrealtime,park2026pimathbfr2reactiverealtimeflow} to enable reactive control under inference latency. A representative example is real-time chunking (RTC)~\cite{black2025realtime, black2025trainingtimeactionconditioningefficient}, which uses action inpainting to generate the next action chunk while the current one is still executing. While such methods enable smooth inference despite latency, they offer no natural mechanism for moving beyond the training data distribution. While methods like RTC can be used for reinforcement learning as a base policy, directly applying RL leaves performance on the table since actions are predicted from previous observations, and thus cannot fully capture the reliability gains that reinforcement learning affords by improving on the base policy. This gap motivates our work of enabling reinforcement learning fine-tuning for real-time policies, so as to simultaneously obtain the reliability benefits of reinforcement learning and the reactivity benefits of real-time execution. 

Our key insight is that reinforcement learning can allow the VLA policy to be maximally reactive by editing the action to be executed using the latest observation at the time of execution, with the base action being generated from the VLA using previous observations given the inference latency. We build on EXPO-FT~\citep{dong2026expoftsampleefficientreinforcementlearning}, a system for sample-efficient, reliable VLA fine-tuning with reinforcement learning that features a large VLA base policy and a small edit policy. Specifically, we turn pretrained VLA policies into reactive controllers by decoupling slow action generation from fast, observation-conditioned action generation.  We assign the base VLA, a large policy with a strong behavior prior, to be responsible for initial action chunk generation, which can incur substantial inference delay, and the edit policy to perform fast, reactive decision-making by editing actions in response to changes in the observation. Concretely, at inference, the VLA proposes candidates of action chunks; once the actions have been generated and a delay has elapsed, the edit policy transforms the remaining actions taking into account the latest observation and selects the chunk of remaining actions with the highest Q-value. We instantiate these ideas as \methodname, a reinforcement learning (RL) framework for fine-tuning real-time VLA policies.

Our key contribution is a framework for reinforcement learning fine-tuning of real-time policies. We evaluate our approach in both simulation and the real world on highly dynamic and stochastic tasks that require both fast reaction and accurate control. In simulation, we evaluate on the Kinetix benchmark \cite{matthews2025kinetix} following prior work on real-time control, and empirically show that \methodname enables a delayed policy to achieve the best performance among delayed and non-delayed methods in 10 out of 10 environments. In the real world, we evaluate on four dynamic manipulation tasks, including robot object passing, ball balancing, table soccer kicking, and dynamic object picking. These tasks involve rapid state changes, stochastic outcomes, and tight timing requirements. Across these tasks, capping the online robot data to 10 minutes, our method improves the average evaluation policy performance from \(42\%\) (12.5/30) to \(97\%\) (29/30), all without human intervention during training, demonstrating rapid and sample-efficient adaptation to challenging real-world dynamics.

\section{Related Work}

\icra{\noindent \textbf{Deep Reinforcement Learning for Robotic Manipulation and Vision-Language-Action Models.} Reinforcement learning has been widely used to improve manipulation policies~\cite{levine2016endtoendtrainingdeepvisuomotor,zhu2018dexterousmanipulationdeepreinforcement,haarnoja2019softactorcriticalgorithmsapplications,mandlekar2020irisimplicitreinforcementinteraction,sharma2023selfimprovingrobotsendtoendautonomous,drloffpolicy,luo2025serlsoftwaresuitesampleefficient,ankile2025residualoffpolicyrlfinetuning,lei2026rl100performantroboticmanipulation}. These methods typically optimize lightweight Gaussian policies, which provide low inference latency and support high-frequency control for dynamic manipulation tasks. However, their limited policy capacity causes them to often require substantial numbers of samples and be limited to a small breadth of initial states. 
Recent work has explored reinforcement learning for vision-language-action models. However, modern VLAs \cite{intelligence2025pi05visionlanguageactionmodelopenworld,nvidia2025gr00tn1openfoundation,geminiroboticsteam2025geminirobotics15pushing,intelligence2025pi06vlalearnsexperience,black2026pi0visionlanguageactionflowmodel} often employ expressive generative policies, such as diffusion or flow-matching policies, making conventional continuous-control RL algorithms difficult to apply directly. Several works develop RL algorithms tailored to diffusion or flow-based policies \cite{mark2024policyagnosticrloffline,ren2024diffusionpolicypolicyoptimization,psenka2025learningdiffusionmodelpolicy,zhang2026reinflowfinetuningflowmatching,patil2026ogposampleefficientfullfinetuning,dong2026tqlscalingqfunctionstransformers,dong2026valueflows,li2026qlearningadjointmatching,dong2026fastervalueguidedsamplingfast,dong2026expostablereinforcementlearning}. Multiple online RL methods use on-policy algorithms to finetune
the VLA~\cite{ren2024diffusionpolicypolicyoptimization,zhang2026reinflowfinetuningflowmatching,chen2026pitextttrlonlinerlfinetuning}, which can require extensive environment interaction. Consequently, recent work has increasingly explored off-policy RL for VLA finetuning \cite{wagenmaker2025steering,lu2025vlarlmasterfulgeneralrobotic,xiao2025selfimprovingvisionlanguageactionmodelsdata,chen2025conrftreinforcedfinetuningmethod,xu2026rltokenbootstrappingonline,dong2026expoftsampleefficientreinforcementlearning}, as it can reuse previously collected experience and substantially improve sample efficiency.
Our work builds on  EXPO-FT~\cite{dong2026expoftsampleefficientreinforcementlearning}, a system for sample-efficient, reliable finetuning of VLA models
with reinforcement learning, and we focus on a largely unexplored challenge: enabling RL fine-tuning while meeting the real-time control requirements of dynamic real-world manipulation. In particular, whereas prior works have shown improving policy performance and sample efficiency from RL fine-tuning, we study how to obtain these benefits while achieving sufficiently low inference latency for high frequency control in dynamic environments. 

\noindent \textbf{Real-Time Inference for Vision-Language-Action Models.} VLA models typically incur substantial inference latency due to their large size, limiting their ability to support high-frequency control. Prior works have explored system-level techniques to accelerate VLA inference~\citep{yang2025efficientvla,ma2025runningvlasrealtimespeed,lu2026faster,niu2026realtimevlaflashspeculativeinference}, including model compression~\citep{yang2025efficientvla}, kernel-level optimization~\citep{ma2025runningvlasrealtimespeed}, accelerated sampling~\citep{lu2026faster}, and speculative inference~\citep{niu2026realtimevlaflashspeculativeinference}. Another line of work modifies the inference or training procedure to enable more responsive execution. One strong, widely used approach is real-time chunking (RTC)~\cite{black2025realtime}, which uses action inpainting to generate the next action chunk while the current chunk is still being executed, thereby enabling asynchronous VLA execution. Subsequent works incorporate this capability directly into policy training~\citep{black2025trainingtimeactionconditioningefficient,tang2026vlashrealtimevlasfuturestateaware}. $\pi\mathbf{R}^2$~\citep{park2026pimathbfr2reactiverealtimeflow} combines a fast proprioceptive channel alongside a slow updated vision-language channel. Other approaches use auxiliary modules to refine actions between successive VLA updates~\cite{sendai2025leaveobservationbehindrealtime,jiang2026futurertcrealtimerobotexecution}. Different to these approaches, we focus on the reinforcement learning setting with the goal of enabling reinforcement learning for real-time
policies. These prior methods are general and can in principle be combined with RL post-training; our insight, however, is that the structure of modern RL algorithms enables us to design an approach that is even more performant (\Cref{fig:sim_exp} and \Cref{fig:real_exp}).

}

\arxiv{\noindent \textbf{Deep Reinforcement Learning for Robotic Manipulation.} Reinforcement learning has been widely used to improve manipulation policies through direct interaction with the environment \cite{levine2016endtoendtrainingdeepvisuomotor,zhu2018dexterousmanipulationdeepreinforcement,haarnoja2019softactorcriticalgorithmsapplications,mandlekar2020irisimplicitreinforcementinteraction,sharma2023selfimprovingrobotsendtoendautonomous,drloffpolicy,luo2025serlsoftwaresuitesampleefficient,ankile2025residualoffpolicyrlfinetuning,lei2026rl100performantroboticmanipulation}. However, real-world interaction is costly, making sample efficiency a fundamental challenge. Prior work addresses this challenge through algorithmic design \cite{haarnoja2018soft,chen2021randomizedensembleddoubleqlearning,ball2023efficientonlinereinforcementlearning,nauman2024biggerregularizedoptimisticscaling,dong2025reinforcementlearningimplicitimitation,luo2025precisedexterousroboticmanipulation} to enable effective policy improvement from limited experience. These methods typically optimize lightweight Gaussian policies, which provide low inference latency and support high-frequency control for dynamic manipulation tasks. However, their limited policy capacity prevents them from leveraging the broad behavioral priors of large pretrained models, often requiring substantial number of samples and small breadth of initial states. Our work bridges this gap by enabling sample-efficient RL adaptation of pretrained VLAs while satisfying the latency and reactivity constraints of real-time manipulation.

\noindent \textbf{Reinforcement Learning for Vision-Language-Action Models.} 
Recent work has explored reinforcement learning  for finetuning pretrained vision-language-action  models. However, modern VLAs \cite{intelligence2025pi05visionlanguageactionmodelopenworld,nvidia2025gr00tn1openfoundation,geminiroboticsteam2025geminirobotics15pushing,intelligence2025pi06vlalearnsexperience,black2026pi0visionlanguageactionflowmodel} often employ expressive generative policies, such as diffusion or flow-matching policies, making conventional continuous-control RL algorithms difficult to apply directly. Several works address this mismatch by developing RL algorithms tailored to diffusion or flow-based policies \cite{mark2024policyagnosticrloffline,ren2024diffusionpolicypolicyoptimization,psenka2025learningdiffusionmodelpolicy,zhang2026reinflowfinetuningflowmatching,patil2026ogposampleefficientfullfinetuning,dong2026valueflows,li2026qlearningadjointmatching,dong2026qlearningworldmodels,dong2026fastervalueguidedsamplingfast,dong2026tqlscalingqfunctionstransformers,dong2026expostablereinforcementlearning}. Multiple online RL methods use on-policy algorithms to finetune
the VLA~\cite{ren2024diffusionpolicypolicyoptimization,zhang2026reinflowfinetuningflowmatching,chen2026pitextttrlonlinerlfinetuning}, which can require extensive environment interaction. Consequently, recent work has increasingly explored off-policy RL for VLA finetuning \cite{wagenmaker2025steering,lu2025vlarlmasterfulgeneralrobotic,xiao2025selfimprovingvisionlanguageactionmodelsdata,chen2025conrftreinforcedfinetuningmethod,xu2026rltokenbootstrappingonline,dong2026reallyneedpretrainqfunctions,dong2026expoftsampleefficientreinforcementlearning}, as it can reuse previously collected experience and substantially improve sample efficiency.
Our work builds on  EXPO-FT~\cite{dong2026expoftsampleefficientreinforcementlearning}, a system for sample-efficient, reliable finetuning of VLA models
with reinforcement learning, and we focus on a largely unexplored challenge: enabling RL fine-tuning while meeting the real-time control requirements of dynamic real-world manipulation. In particular, whereas prior work have shown improving policy performance and sample efficiency from RL fine-tuning, we study how to obtain these benefits while achieving sufficiently low inference latency for high frequency control in dynamic environments.

\noindent \textbf{Real-Time Inference for Vision-Language-Action Models.} VLA models typically incur substantial inference latency due to their large size, limiting their ability to support high-frequency control. Prior work has explored system-level techniques to accelerate VLA inference~\citep{yang2025efficientvla,ma2025runningvlasrealtimespeed,lu2026faster,niu2026realtimevlaflashspeculativeinference}, including model compression~\citep{yang2025efficientvla}, kernel-level optimization~\citep{ma2025runningvlasrealtimespeed}, accelerated sampling~\citep{lu2026faster}, and speculative inference~\citep{niu2026realtimevlaflashspeculativeinference}. Another line of work modifies the inference or training procedure to enable more responsive execution. One strong, widely used approach is real-time chunking (RTC)~\cite{black2025realtime}, which uses action inpainting to generate the next action chunk while the current chunk is still being executed, thereby enabling asynchronous VLA execution. Subsequent works incorporate this capability directly into policy training. Training-Time RTC~\citep{black2025trainingtimeactionconditioningefficient} and VLASH~\citep{tang2026vlashrealtimevlasfuturestateaware} modify the standard VLA training procedure by incorporating action conditioning or future-state prediction during training to enable asynchronous action generation during inference. $\pi\mathbf{R}^2$~\citep{park2026pimathbfr2reactiverealtimeflow} combines a fast proprioceptive channel alongside a slow updated vision-language channel. Other approaches use auxiliary policies and correction modules to refine actions between successive VLA updates~\cite{sendai2025leaveobservationbehindrealtime,jiang2026futurertcrealtimerobotexecution}. Different to these approaches, we focus on the reinforcement learning setting with the goal of enabling reinforcement learning fine-tuning for real-time
policies. These prior methods are general and can in principle be combined with RL post-training; our insight, however, is that the structure of modern RL algorithms enables us to design an approach that is even more performant (\Cref{fig:sim_exp} and \Cref{fig:real_exp}).

}

\begin{figure*}[t]
    \centering
    \arxiv{\includegraphics[width=\textwidth]{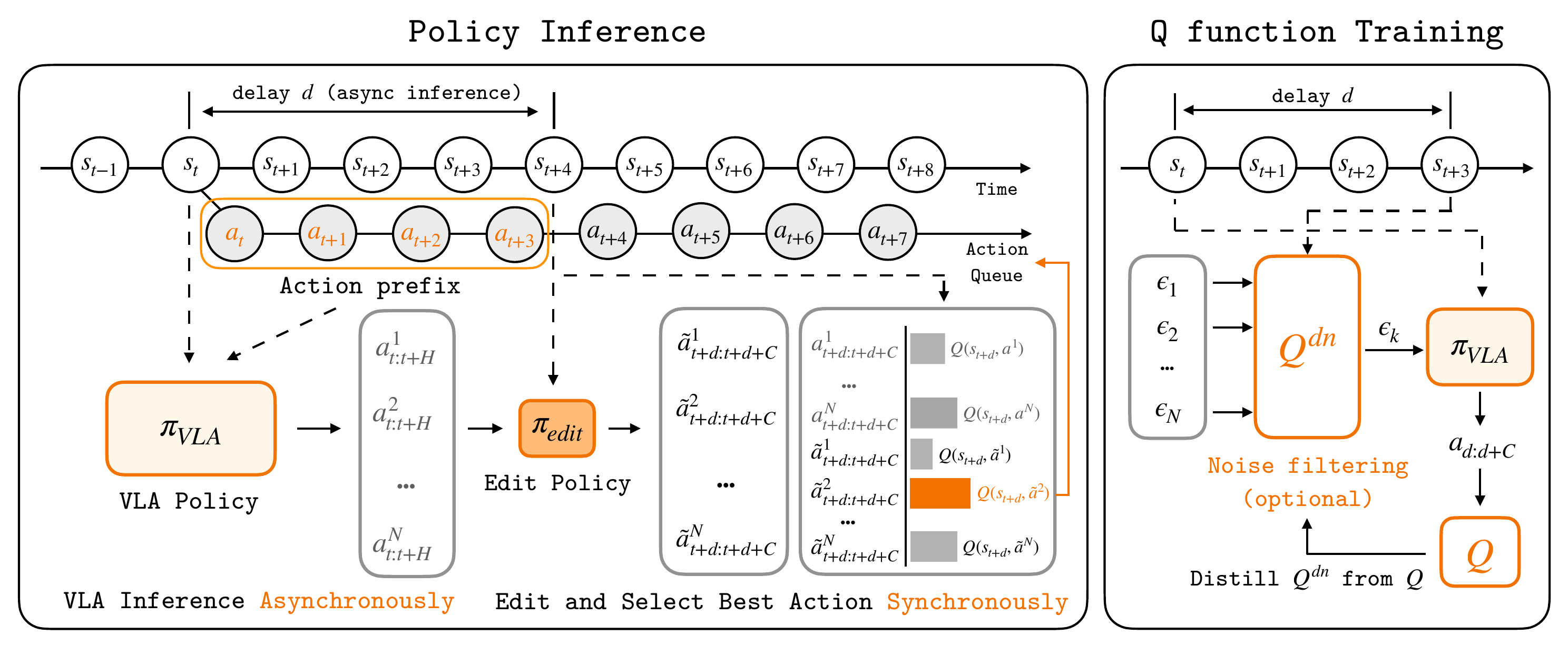}}
    \caption{\textbf{Left: Real-time policy inference of \methodname.} \methodname decouples slow VLA action generation from fast, reactive decision-making. While the robot executes the current action chunk, the VLA asynchronously generates multiple candidate action chunks. Once new actions are required, a lightweight edit policy refines candidates using the latest observation, and a learned Q-function selects the highest-value action chunk for execution.  \textbf{Right: Noise-level filtering during Bellman backup.} During Bellman backup, we filter samples in the noise space to reduce training compute.}
    \label{fig:method}
    \icra{\vspace{-0.5cm}}
\end{figure*}

\section{Background}
We consider the standard reinforcement learning framework, where problems are modeled by a Markov decision process (MDP) $\mathcal{M} = (\mathcal{S}, \mathcal{A}, r, T, \gamma, \rho)$. In the MDP, $\mathcal{S}$ is the state space and $\mathcal{A}$ is the action space. At each timestep $t$, the agent selects action $a_t \in \mathcal{A}$ according to policy $\pi(\cdot \mid s_{t-d}, s_t)$ and receives scalar reward $r(s_t,a_t)\in\mathbb{R}$. The environment transitions following the transition dynamics of the MDP $s_{t+1}\sim T(\cdot\mid s_t,a_t)$, with starting state initialized from $s_0\sim\rho(\cdot)$. The tuple $(s_t,a_t,r_t,s_{t+1})$ is added to the replay buffer $\mathcal{D}$ for learning. The goal of reinforcement learning is to maximize the expected discounted return $\mathbb{E}_{\pi}\!\left[\sum_{t=0}^{T}\gamma^t r(s_t,a_t)\right]$, where $\gamma\in[0,1]$ is the discount factor. We consider the problem of reinforcement learning fine-tuning of VLA models under a real-time constraint, where policy inference time needs to be faster than the control frequency $f$. Because VLAs are large models, inference itself is costly. We denote the delay of timesteps from inference as $d$. Modern VLAs often employs action chunking, predicting a sequence of $H$ future actions $a_{t:t+H}$ at each timestep and executing $C\leq H$ at each timestep. We assume the delay $d$ is less than or equal to execution length $C$. 

\textbf{Training-Time Real-Time Chunking~\citep{black2025trainingtimeactionconditioningefficient}.} When inference incurs a delay of $d$ timesteps, the first $d$ actions of a newly predicted chunk cannot be executed, since by the time they are produced the environment has already advanced to $t+d$. Training-time RTC~\citep{black2025trainingtimeactionconditioningefficient} addresses this issue in the imitation learning setting by additionally executing actions $a_{t+C:t+C+d}$ from the previous chunk while inference for the new chunk is in progress, and conditioning the new prediction on this committed action prefix. Specifically, given the original state $s_{t}$ and the committed action prefix $a^{\text{prev}}_{t:t+d}$, the policy predicts the remaining actions as $a_{t+d:t+d+H} \sim \pi(\cdot \mid s_{t}, a^{\text{prev}}_{t:t+d})$. By conditioning on both the current state and the already-committed actions, the policy can account for inference latency and produce a coherent continuation of the action chunk. During training, the delay $d$ is randomly sampled across a range of values so that the policy learns to remain robust to varying inference latencies at deployment time.

\textbf{EXPO and EXPO-FT~\cite{dong2026expostablereinforcementlearning,dong2026expoftsampleefficientreinforcementlearning}. } To finetune the VLA policy with RL, we build on EXPO~\citep{dong2026expostablereinforcementlearning}, a recently proposed RL algorithm that is both highly sample-efficient and stable for training expressive policies. Classical sample-efficient RL algorithms are designed around Gaussian policies and cannot be directly applied to pretrained VLAs, which typically use flow or diffusion policies; EXPO instead provides a principled foundation for RL fine-tuning in this regime.

EXPO couples two parameterized policies. The first is a base flow policy---in our setting the VLA model $\pi_\text{VLA}$, obtained through supervised training---and the second is a small edit policy $\pi_\text{edit}$ whose objective is to maximize the learned Q-value:

{\small
\begin{equation}
\begin{aligned}
\mathcal{L}(\pi_\text{edit})
= -\mathbb{E}_{(s_t,a_t)\sim\mathcal{D},\;\hat{a}_t\sim\pi_\text{edit}}
\bigl[
&Q_\phi(s_t, a_t + \hat{a}_t) \\
&-\alpha \log \pi_\text{edit}(\hat{a}_t|s_t,a_t)
\bigr]
\end{aligned}
\end{equation}
}
Rather than altering the base action outright, $\pi_\text{edit}$ predicts a bounded edit $\hat{a}$ restricted to $[-\beta, \beta]$, which is summed with the base action $a$ to yield the edited action $\tilde{a} = a + \hat{a}$, transforming the base action to a higher value distribution. Confining the Q-function signal to this edit avoids backpropagation of the Q-value to the VLA backbone and also anchored to actions already close to optimal. During rollout or backup target construction, EXPO uses an on-the-fly (OTF) policy that chooses the highest-value action candidate from the base and edited actions:

{\small
\begin{equation}
\tilde{a}^* = \underset{a \;\in\; \bigcup_{i=1}^{N}\{a_i,\,\tilde{a}_i\}}{\arg\max}\; Q_\phi(s, a)
\end{equation}}
The critic itself is fit by temporal-difference:
{\small
\begin{equation}
\begin{aligned}
\mathcal{L}(\phi)
= \mathbb{E}_{(s_t,a_t,s_{t+1})\sim\mathcal{D}}
\Bigl[
\bigl(
r_t + \gamma
&Q_{\phi'}(s_{t+1}, \tilde{a}^{*}_{t+1}) \\
&- Q_{\phi}(s_t,a_t)
\bigr)^2
\Bigr]
\end{aligned}
\end{equation}
}

EXPO-FT~\citep{dong2026expoftsampleefficientreinforcementlearning} is a system on top of EXPO to finetune VLA models, incorporating human-in-the-loop and action chunking. We build directly on top of EXPO-FT. While human intervention can provide useful corrective signals, we do not use human intervention for the experiments in this paper.

\section{\methodname}
In this section, we present a complete framework for fine-tuning VLA models with reinforcement learning for real-time control. Our goal is to efficiently enable pretrained VLA policies to reach high reliability in dynamic environments. We first formalize the learning setting and objective (Section \ref{method:problem}), then introduce our real-time RL algorithm for VLA models (Section \ref{method:algo}), and finally describe the training procedure (Section \ref{method:procedure}).

\subsection{Problem Statement} \label{method:problem}
We consider the problem of finetuning vision-language-action models $\pi_{\text{VLA}}$ using reinforcement learning for real-time robotic control. Policy inference, especially for a large policy, may take non-negligible time, such that naively executing the predicted actions with a delay will result in distribution shift and the action executed at time $t+d$ has to be computed from earlier observations to mitigate the compute latency.

\arxiv{A common simplifying assumption in prior work is that policy inference is effectively instantaneous, so the action computed from an observation can simply be executed once inference finishes. This assumption is increasingly untenable for large VLA policies, and it is especially costly to make during RL fine-tuning because standard RL assumes the Markov property, that the executed action is a function of the current state, $a_t \sim \pi(\cdot \mid s_t)$; under latency the action at $t$ is actually a function of $s_{t-d}$, so the delayed process is no longer Markovian in $s_t$, and applying standard RL updates as if can bias credit assignment. This is specifically a problem when running RL for high reliability, where actions are refined precisely. Treating inference latency as negligible therefore risks significantly undermining the reliability that RL fine-tuning can deliver, motivating the need for an explicit solution.  }  

For tasks that require real-time execution, we empirically observe that existing $\pi_{\text{VLA}}$ models cannot achieve satisfactory success rates out of the box. Therefore, following the standard online RL finetuning practices, we assume access to a small offline dataset of expert demonstrations, $\mathcal{D}_0$, collected via either human teleoperation or scripted policies operating at the same control frequency. We adopt a sparse binary reward $r \in [0,1]$ indicating successful task completion. The task completion classifier may be either rule-based or learned. Observations consist of multi-view RGB images from a wrist-mounted camera and a fixed side-view camera, augmented with the robot’s proprioceptive state. Policy parameters are updated either after every environment step, at the end of each episode, or at fixed episode-batch intervals. The objective is to maximize the task success rate.

To address the latency of $\pi_{\text{VLA}}$ under high-frequency control, rather than optimizing the inference latency of the VLA model itself, we focus on asynchronously fine-tuning and executing the policy while maintaining a fixed real-time control frequency. Suppose the VLA model requires $t$ seconds for each inference and the robot operates at a control frequency of $f$ Hz. The inference process therefore incurs a delay of approximately $d = \lfloor t\times f \rfloor + 1$ control steps, meaning that to execute a new action at timestep $t+d$, inference must be initiated approximately $d$ control steps earlier, at timestep $t$. For typical hardware and VLA models, we assume $1 \leq d \leq C$, where $C$ denotes the execution horizon of the policy. This inference delay creates a mismatch between the observation used to initiate VLA inference and the robot state at which the resulting action is eventually executed. Our goal is therefore to fine-tune the VLA policy using RL to explicitly account for this delay and improve policy performance under real-time control. In the following section, we discuss how
these challenges are addressed in our approach, \methodname.

\subsection{RL for Real-Time Vision-Language-Action Policies} \label{method:algo}

We build on EXPO-FT~\cite{dong2026expoftsampleefficientreinforcementlearning} as a sample efficient RL fine-tuning framework. \methodname addresses the real-time inference challenge by decoupling action generation into two timescales: a slow, asynchronous step that generates candidate action chunks ahead of execution, and a fast, synchronous step that edits and selects among the highest value candidates at the time of execution, conditioned on the most recent observation. The method is illustrated in \Cref{fig:method}.

\noindent \textbf{Asynchronous VLA Action Generation.} The VLA is a large, expressive model for capturing behaviors, and because of its scale, calling the model during inference introduces a delay of $d$ steps. We therefore begin inference from $\pi_{\mathrm{VLA}}$ asynchronously at time $t$ when $d$ steps remain in the currently queued action chunk. We sample multiple candidate action chunks from the observation $s_{t}$. Following the training-time RTC formulation \cite{black2025trainingtimeactionconditioningefficient}, the actions from the previous chunk, $a^{\text{prev}}_{t:t+d}$, that are executed during the VLA inference window are inpainted as conditioning input when sampling future actions. Specifically, for each candidate $i$, we sample:\arxiv{

}
\begin{equation}
\label{eq}
a^i_{t:t+H} = \pi_{\mathrm{VLA}}(s_{t}, a^{\text{prev}}_{t:t+d}, \epsilon^i),
\qquad
\epsilon^i \sim p(\epsilon)
\end{equation}

where $\epsilon^i$ denotes the sampling noise used to produce diverse VLA candidates. Since the first $d$ actions correspond to the inference-delay window, we retain only the subsequent $C$-step action segment $ a^i_{t+d:t+d+C}$ for each action candidate. 

\noindent \textbf{Fast, Synchronous Edits.} 
Once the environment arrives at the latest observation $s_{t+d}$ for execution, a lightweight edit policy $\pi_{\theta}^{\mathrm{edit}}$ transforms each candidate based on the latest observation to account for state changes during the inference delay and maintain reactivity:

\begin{equation}
\hat{a}^i_{t+d:t+d+C}
\sim
\pi_{\theta}^{\mathrm{edit}}
\left(
\cdot \mid s_{t+d}, a^i_{t+d:t+d+C}
\right)
\end{equation}

The edited actions are $\tilde{a}^i_{t+d:t+d+C}=a^i_{t+d:t+d+C}+\hat{a}^i_{t+d:t+d+C}$. Finally, the action critic $Q_{\phi}$ evaluates both the original and edited candidates under the current state and selects the highest-value action chunk for execution:
\begin{equation}
\tilde{a}^*_{t+d:t+d+C}
=
\underset{
a \in
\bigcup_{i=1}^{N}
\{a^i_{t+d:t+d+C},\tilde{a}^i_{t+d:t+d+C}\}
}{\arg\max}
\;
Q_{\phi}(s_{t+d},a)
\end{equation}
This design enables expensive VLA inference to run asynchronously in the background while preserving fast, state-aware correction and selection immediately before execution.

\noindent \textbf{Training Objective. }
We now describe how the action critic $Q_{\phi}$, edit policy $\pi_{\mathrm{edit}}$, and VLA $\pi_{\mathrm{VLA}}$ are updated. We use $Q_{\phi'}$ to denote the target critic.

We fine-tune $\pi_{\mathrm{VLA}}$ with the Training-Time RTC objective \cite{black2025trainingtimeactionconditioningefficient} on both offline demos and online data from rollouts. Following RTC, we simulate inference delay during training by splitting each ground-truth action chunk into a $d$-step action prefix and a remaining action postfix. The prefix is provided to the policy as clean, non-noisy actions with its flow-matching timesteps set to $1$, while noise is added only to the postfix. The flow-matching loss is masked to the postfix:

{\small
\begin{equation}\label{eq:bc}
\begin{aligned}
    A_t^\tau
    &= \tau A_t + (1-\tau)\epsilon,
    \qquad
    \epsilon \sim \mathcal{N}(0,I), \\
    \mathcal{L}_{\mathrm{BC}}(\pi_{\mathrm{VLA}})
    &=
    \mathbb{E}_{(s_t,A_t)\sim\mathcal{D}}
    \Big[
    \big\|
    \mathbf{m}_d \odot
    \big(
    v_{\mathrm{VLA}}(A_t^\tau,s_t,\boldsymbol{\tau}_d)
    \\
    &\hspace{3.55cm}
    -(\epsilon-A_t)
    \big)
    \big\|_2^2
    \Big]
\end{aligned}
\end{equation}
}

where $\mathbf{m}_d$ masks out the first $d$ actions from the loss and $\boldsymbol{\tau}_d$ assigns flow-matching timestep $1$ to the prefix and $\tau$ to the postfix. We use the specific execution delay $d$ for the state after first chunk, and $0$ for the first $[0, C]$ state in each episode during training.

\begin{figure*}[t]
\centering
\includegraphics[width=\icra{0.85}\textwidth]{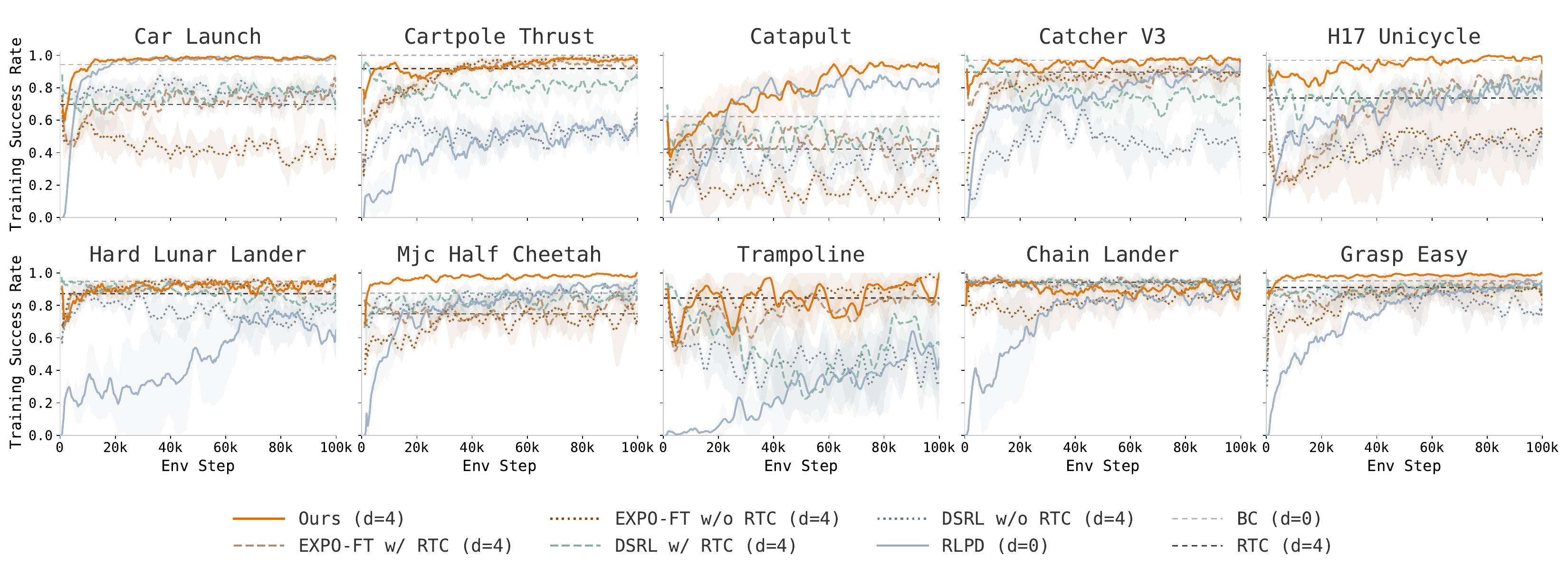}
\caption{\textbf{Training success across all 10 Kinetix \cite{matthews2025kinetix} simulation tasks.} The vertical dashed lines indicate the evaluation performance of two BC baselines, one with no inference delay and one with a 4-step delay. Each setting is run with 4 seeds.}
\icra{\vspace{-0.5cm}}  
\label{fig:sim_exp}
\end{figure*}

We train the edit policy to bring the base actions from the VLA toward higher value regions like in EXPO~\citep{dong2026expostablereinforcementlearning} and EXPO-FT~\citep{dong2026expoftsampleefficientreinforcementlearning}: \arxiv{

}
{\small
\begin{equation}
\begin{aligned}
\mathcal{L}(\pi_\text{edit}) = -\mathbb{E}&_{\substack{(s_t,a_{t:t+C})\sim\mathcal{D}\\[1pt]
\hat{a}_{t:t+C}\sim\pi_\text{edit}}}
\Big[Q_{\phi}\big(s_t, a_{t:t+C} + \hat{a}_{t:t+C}\big) \\
&-\alpha \log \pi_\text{edit}\big(\hat{a}_{t:t+C}\mid s_t,a_{t:t+C}\big)
\Big]
\end{aligned}
\end{equation}
}

Because the edit is conditioned on the latest observation, the combined base-plus-edit policy remains Markovian with delays. 
We train the critic to fit a chunk-level temporal-difference backup, in which one transition spans the full execution horizon $C$ and the bootstrap term is evaluated at the next chunk $\tilde{a}^{*}_{t+C:t+2C}$,\arxiv{

}
{\small
\begin{equation}
\begin{aligned}
\mathcal{L}(Q_{\phi})
= \mathbb{E}_{(s_t,a_{t:t+C},s_{t+C})\sim\mathcal{D}}
\Big[
\big(
r_t
&+ \gamma^{} Q_{\phi'}\big(s_{t+C}, \tilde{a}^{*}_{t+C:t+2C}\big) \\
&- Q_{\phi}\big(s_t,a_{t:t+C}\big)
\big)^2
\Big]
\end{aligned}
\end{equation}
}

Constructing the next chunk $\tilde{a}^{*}_{t+C:t+2C}$ is computationally expensive, as naively sampling \(N\) candidates from \(\pi_{\mathrm{VLA}}\) requires \(N\) full VLA rollouts. We instead (optionally) search in noise space using a lightweight critic \(Q^{\mathrm{dn}}_\psi\), which scores candidate noise without denoising full actions, following FASTER \cite{dong2026fastervalueguidedsamplingfast}. At backup time, we select the best noise, decode it once with \(\pi_{\mathrm{VLA}}\) to obtain a, generate \(K\) edits with \(\pi_{\mathrm{edit}}\), and use the target critic to select the final action chunk,

\begin{equation} \label{eq:noise}
\begin{aligned}
\epsilon^*
&= \arg\max_{i \in [N]}
Q^{\mathrm{dn}}_\psi\big(s_{t+C-d}, \epsilon_i\big), \\
a
&= \Big[
\pi_{\mathrm{VLA}}\big(
s_{t+C-d},
a^{\mathrm{prev}}_{t+C-d:t+C},
\epsilon^*
\big)
\Big]_{d:d+C}, \\
\tilde{a}^{*}_{t+C:t+2C}
&= \operatorname*{arg\,max}_{a' \in \{a,\,a+\hat{a}\}}
Q_{\phi'}\big(s_{t+C}, a'\big)
\end{aligned}
\end{equation}

where $\epsilon_i \sim \mathcal{N}(0,I)$, $\hat{a} \sim \pi_{\mathrm{edit}}(\cdot \mid s_{t+C},a)$. In this way the candidate set is filtered once in noise space and once again after editing, while only a single VLA decode is required per backup regardless of $N$. Finally, $Q^{\mathrm{dn}}_\psi$ is kept consistent with the action critic by regressing it onto the value that $Q_{\phi'}$ assigns to the decoded chunk, with a stop-gradient on the target,\arxiv{

}
{\small
\begin{equation} 
\begin{aligned}
\mathcal{L}(Q^{\mathrm{dn}}_\psi)
= \mathbb{E}\Big[
\big(
&Q^{\mathrm{dn}}_\psi\big(s_{t+C}, \epsilon^{*}\big) \\
&- \operatorname{sg}\!\big[Q_{\phi'}\big(s_{t+C}, \tilde{a}^{*}_{t+C:t+2C}\big)\big]
\big)^2
\Big]
\end{aligned}
\end{equation}
}

so that noise-space selection inherits the ranking of the action-space critic as the latter improves. This allows a large number of action candidates to be used during training, which is helpful for accelerating training.

\subsection{Implementation Details} \label{method:procedure}

\noindent \textbf{Vision Backbone.} At each timestep $t$, the state $s_t$ comprises visual observations from the side and wrist cameras and robot joint positions, and is provided to both actor and critic. VLA actor processes the visual observations using its pretrained visual encoder. Following EXPO-FT~\citep{dong2026expoftsampleefficientreinforcementlearning}, we equip critic with a separate, lightweight ResNet-50 encoder, which achieves high task performance at low computational cost. \arxiv{Full architectural details are provided in \Cref{sec:app_training}.}

\noindent \textbf{Training Procedure. }
For each task, we begin by configuring the camera setup and defining the reward signal, which can take the form of a rule-based criterion or a learned binary classifier. Before online RL begins, we collect a set of demonstrations and fine-tune the VLA using imitation learning with \Cref{eq:bc} under a randomized delay $d$, until it reaches a success rate of around 30\% or higher. This data is then used to initialize the replay buffer. Starting from the supervised fine-tuned VLA policy, we then begin online RL training. The actor executes rollouts in the environment without human intervention, and the policy is updated after each episode, depending on the task requirements and available computational resources.

\noindent \textbf{Reward Classifier.} Reliable deployment requires an accurate and robust reward signal. To minimize task-specific reward engineering, we use a sparse binary reward for all tasks. Specifically, we define a rule-based classifier for each task that assigns a positive reward only when the task is successfully completed, and zero otherwise. \arxiv{The reward classifiers for all tasks are detailed in \Cref{appendix:task_setting}.} This simple formulation avoids dense reward design while remaining effective across a diverse set of tasks.

\section{Experiments}
We evaluate \methodname on ten environments in the Kinetix environment \cite{matthews2025kinetix} and four dynamic real-world tasks, comparing against strong prior methods.  

\subsection{Baselines}

\textbf{RLPD \cite{ball2023efficientonlinereinforcementlearning}.} RLPD is a sample-efficient off-policy reinforcement learning algorithm based on Soft Actor-Critic (SAC) \cite{haarnoja2018soft}, using balanced sampling between offline demonstrations and online experience. It has demonstrated strong performance across robotic manipulation tasks. RLPD uses a lightweight Gaussian policy as the actor, enabling efficient high-frequency control. Since it is naturally suited for real-time control, we run it as is without modification.

\textbf{DSRL \cite{wagenmaker2025steering}, DSRL w/ RTC.} DSRL is a recent reinforcement learning method for finetuning pretrained diffusion and flow-matching policies. Rather than directly optimizing the policy weights, it learns to predict noise input to the pretrained policy. However, applying DSRL directly to high-frequency control, such as 30 Hz, causes the policy to pause between action chunks, disrupting real-time execution. We therefore also evaluate DSRL with training-time RTC \cite{black2025trainingtimeactionconditioningefficient}.

\textbf{EXPO-FT \cite{dong2026expoftsampleefficientreinforcementlearning}, EXPO-FT w/ RTC.} EXPO-FT is a state-of-the-art reinforcement learning framework for finetuning pretrained VLA policies. Similar to \methodname, it adopts EXPO \cite{dong2026expostablereinforcementlearning} for policy improvement. However, directly applying EXPO-FT to high-frequency control can cause pauses between action chunks at 30 Hz. We therefore also evaluate EXPO-FT with training-time RTC \cite{black2025trainingtimeactionconditioningefficient} for a stronger comparison under real-time execution.

\arxiv{
RLPD and DSRL are trained asynchronously, with the learner updating asynchronously at high update-to-data (UTD) ratios; this affords them substantially more gradient updates than EXPO-FT and \methodname, which perform updates per episode. This asymmetry favors RLPD and DSRL. We nonetheless retain this advantage for these prior methods throughout our experiments, as without it their performance degrades considerably, noting that even so, EXPO-FT and \methodname achieve higher performance despite the compute disadvantage.
}

\begin{figure*}[t]
\centering
\icra{\vspace{0.2cm}}
\includegraphics[width=\icra{0.85}\textwidth]{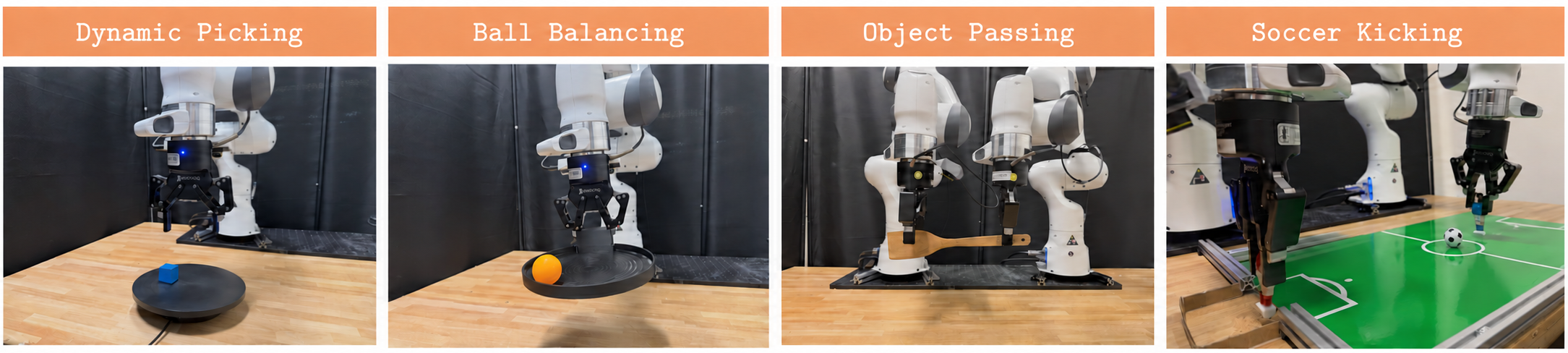}
\caption{\textbf{Four real-world manipulation tasks in our evaluation suite:} Dynamic Picking, Ball Balancing, Object Passing, and Soccer Kicking. All tasks require fast and reactive policies for dynamic environment changes.}
\label{fig:real_result}
\end{figure*}

\subsection{Simulation Experiment}
\textbf{Task Setup.} For our simulation experiments, we evaluate each approach on 10 dynamic tasks from the Kinetix benchmark \cite{matthews2025kinetix}\arxiv{, as shown in \Cref{fig:sim_exp}}, following the RTC setup \cite{black2025realtime}. The environments use force-based control with Gaussian action noise and feature dynamic motions such as catching and balancing, making dynamic control crucial for successful execution. Following the RTC setup \cite{black2025realtime}, we pretrain the RTC flow-matching policy offline on 1M transitions, then finetune it online for 100k environment steps across all tasks. For all RL methods that use a base flow-matching policy, each call to the base policy incurs a 4-step inference delay during online finetuning and evaluation. The policy in RLPD and the edit policy in our method incurs no additional delay due to their lightweight nature, and is evaluated with zero inference delay. As a reference, we additionally report the pretrained BC policy under zero delay, which provides a reference on the performance achievable without inference latency.

\begin{figure*}[t]
\centering
\includegraphics[width=\icra{0.85}\textwidth]{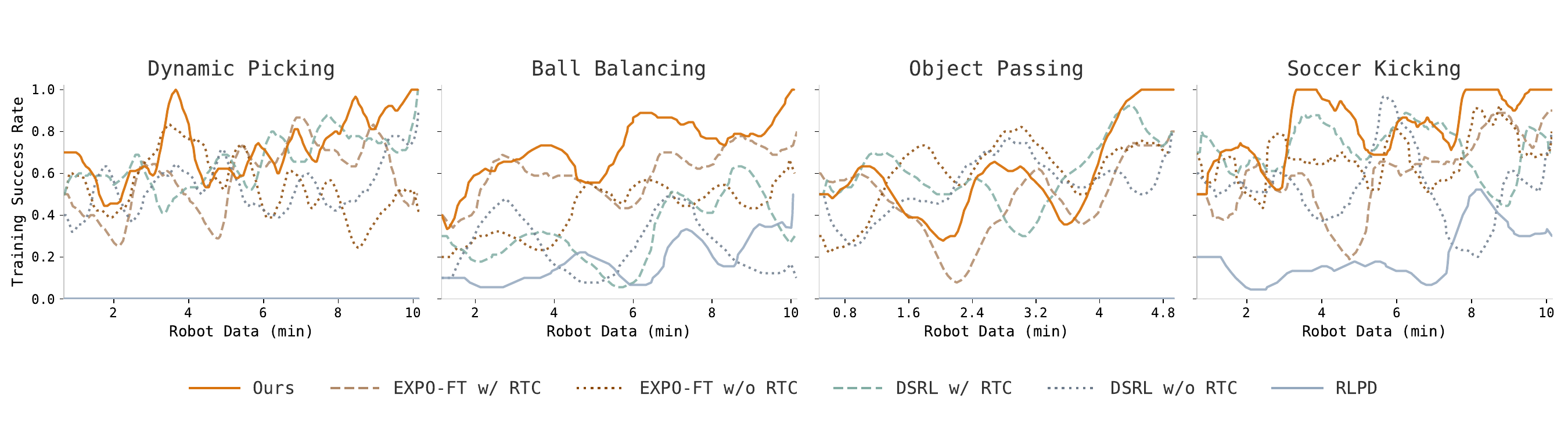}
\caption{\textbf{Training success across four real-world dynamic tasks.} We train each policy with at most 10 minutes of online robot data, or until one of the policies achieves 30/30 success during evaluation.}
\icra{\vspace{-0.1cm}}
\label{fig:real_exp}
\end{figure*}

\begin{table*}[t]
\centering
\setlength{\tabcolsep}{0pt}
\caption{\textbf{Success rates on four real-world tasks.} Each task is evaluated over 30 trials. We train each policy until one policy reaches a 30/30 success rate or 10 minutes of online data have been collected and used for training.}
\label{tab:real_world}
\begin{tabular*}{\icra{0.85}\textwidth}{@{\extracolsep{\fill}}lcccccccc}
\toprule
\multirow{2}{*}{Task} & \multicolumn{8}{c}{Success Rate (x/30). Training is capped at 10 minutes of online data. } \\
\cmidrule(lr){2-9}
& SFT & SFT w/ RTC & RLPD & DSRL & DSRL w/ RTC & EXPO-FT & EXPO-FT w/ RTC & \methodname \\
\midrule
Dynamic Picking
& 19/30 & 22/30 & 0/30 & 23/30 & 25/30 & 21/30 & 24/30 & \textbf{30/30} \\

Ball Balancing
& 8/30 & 12/30 & 12/30 & 11/30 & 15/30 & 18/30 & 23/30 & \textbf{28/30} (10 min reached)\\

Object Passing
& 10/30 & 22/30 & 0/30 & 23/30 & 23/30 & 19/30 & 27/30 & \textbf{30/30} \\

Soccer Kicking
& 13/30 & 16/30 & 6/30 & 16/30 & 17/30 & 17/30 & 26/30 & \textbf{28/30} (10 min reached)\\

\midrule
Average
& 12.5/30 & 18/30 & 4.5/30 & 18.3/30 & 20/30 & 18.8/30 & 25/30 & \textbf{29/30} \\
\bottomrule
\end{tabular*}
\icra{\vspace{-0.5cm}}  
\end{table*}

\textbf{Experiment Results.} Now we present the simulation results. \Cref{fig:sim_exp} shows the training curves\arxiv{, while the full evaluation results are reported in \Cref{sec:app_sim_results}}. \methodname achieves an average success rate of $96.2\%$ under the 4-step delay, substantially outperforming all delayed RL baselines. In particular, it improves over DSRL, DSRL w/ RTC, EXPO-FT, and EXPO-FT w/ RTC by $34.5$, $20.1$, $21.3$, and $14.5$ percentage points, respectively, demonstrating that explicitly addressing stale observations and delayed action generation is broadly effective across dynamic tasks and delayed settings. Importantly, \methodname not only compensates for inference delay but also surpasses the no-delay RLPD baseline on average. While the RLPD policy achieves an average success rate of $81.4\%$ when evaluated with zero delay, \methodname reaches $96.2\%$ while operating with a 4-step inference delay for the base flow-matching policy and no delay for only the action edit. This result is particularly notable because the RLPD policy has access to the current observation when producing each action, whereas \methodname must generate the candidate actions based on delayed observations. \arxiv{We present additional experiments on the effectiveness of noise filtering evaluated in simulation in \Cref{sec:app_value}.}

\subsection{Real-World Experiment}
\noindent\textbf{Task Setup.} In all real-world experiments, the robot is controlled in end-effector space using Cartesian and gripper velocity commands at 30 Hz. At each timestep, the policy receives two $224 \times 224$ RGB images from side- and wrist-mounted cameras, along with proprioceptive observations comprising the end-effector position and orientation. Environment resets are performed either automatically or by a human operator, depending on the task. We evaluate \methodname on four real-world tasks, shown in \Cref{fig:real_result}. 

\noindent \textit{Ball Balancing. } The Ball Balancing task requires the robot to control a black plate with a ping-pong ball on top. The robot must continuously rotate the plate to keep the ball near its center for several consecutive frames. The task is highly dynamic and stochastic, requiring smooth and reactive control to small changes in the ball's position.

\noindent\textit{Dynamic Picking. } The Dynamic Picking task requires the robot to pick up a block placed on a rotating plate. The block can start at arbitrary positions, resulting in different motion speeds and trajectories. The policy must infer the block's motion and quickly reach and grasp it before it moves away.

\noindent\textit{Object Passing. } The Object Passing task requires the robot to receive an object from another robot arm with random motion. The policy must continuously react to the motion of the other arm and rapidly move the end effector toward the object's predicted position. The unpredictability of the object's motion makes timely and reactive control essential.

\noindent\textit{Soccer Kicking. } The Soccer Kicking task requires the robot to kick a ball into a small goal while avoiding a moving defender. Successful execution requires precise spatial control and timing. The policy must determine both where to kick the ball and when to initiate the kick.

The VLA policy has an inference latency of approximately 67 ms on our server. For Ball Balancing, Object Passing, and Soccer Kicking, we introduce an additional 100 ms delay to simulate the higher inference latency associated with more constrained computing resources or a larger model, yielding a total latency of approximately 167 ms. For these three tasks, we set the delay $d$ to 5, corresponding to approximately 167 ms. For Dynamic Picking, we retain the original inference latency of 67 ms and set the delay $d$ to 3. Adding further latency reduces the success rates of all non-asynchronous baselines to nearly zero, as delayed gripper closure prevents timely grasping of the moving block.

Given the number of baselines and the computational cost, \uline{we cap training at 10 minutes of online robot interaction} per task for all methods. We evaluate each task over 30 trials. Success is independently verified by a human observer. \arxiv{Further details on reward definitions, success detection, reset procedures, and task randomization are provided in \Cref{appendix:task_setting}.}

\textbf{Experiment Results.} We now present our experimental results on the four dynamic real-world tasks. As shown in \Cref{fig:real_exp} and \Cref{tab:real_world}, \methodname consistently achieves near-perfect performance across all four tasks, with an average success rate of 29/30, substantially outperforming all prior methods. In particular, Ball Balancing exhibits substantial environmental randomness, where small perturbations in the ball's motion can lead to significantly different future states, making rapid adaptation to the latest observation particularly important. On this task, \methodname achieves 28/30 success, while no prior method exceeds 23/30. Overall, \methodname significantly improves over prior methods. 

\arxiv{
\begin{figure}[t]
    \centering
    \includegraphics[width=\columnwidth]{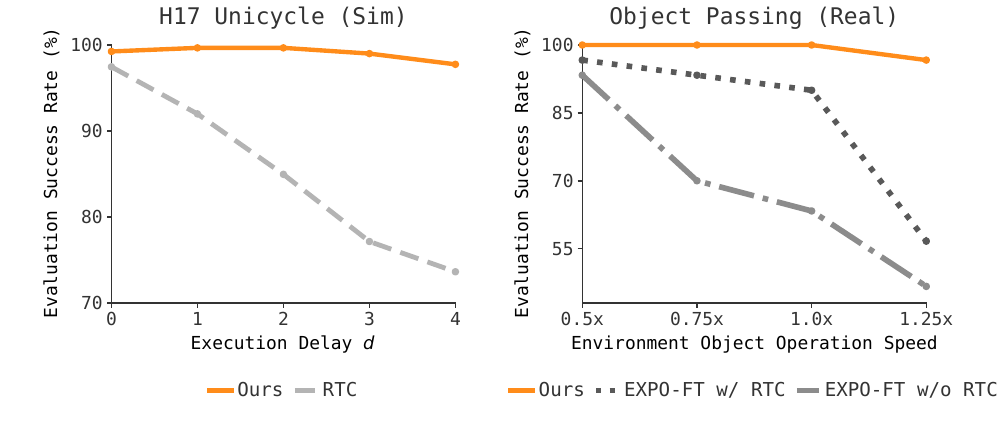}
    \caption{\textbf{Success rates under varying delays and environment speeds.} We evaluate \methodname and the baselines on the H17 Unicycle task with varying delays $d$ and on the Object Passing task with varying passing speeds.}
    \vspace{-0.5cm} 
    \label{fig:speed_analysis}
\end{figure}

\subsection{Performance under varying delays and environment speeds}
To further investigate the effectiveness of \methodname in highly dynamic, real-time settings, we conduct two additional experiments varying inference delays in the H17 Unicycle simulation task and object-passing speeds in the real-world Object Passing task. For varying delays, we train \methodname with different delays to simulate variations in hardware capabilities and base-model inference speeds. As shown in \Cref{fig:speed_analysis}, \methodname maintains stable performance as the delay increases, whereas RTC's performance deteriorates under longer delays. These results highlight the importance of accounting for inference latency in real-time control and demonstrate that our approach can better handle delays. For the object speed experiment, we evaluate \methodname against two EXPO-FT variants across different environment speeds. As shown in \Cref{fig:speed_analysis}, \methodname maintains a near-100\% success rate across all tested speeds, whereas EXPO-FT without real-time degrade in performance as the passing speed increases. These results further demonstrate the importance of fast, reactive action edits for dynamic environments. 
}

\section{Discussion}

We presented \methodname, a framework for RL finetuning of real-time VLA policies. Across a suite of challenging dynamic robotic tasks, \methodname demonstrates rapid sample-efficient adaptation to complex real-world dynamics. Despite these results, \methodname has limitations. First, a human provides environment resets in our experiments, which introduces operational burden; automating the reset process is an important direction for future work. Second, we use task-specific success detector following prior work; however, this requires designing classifier per task. While we do not explore alternative reward specifications in this work, identifying which reward formulation performs best remains an open question for future work.

\arxiv{
\section{Acknowledgments}

This work was in part supported by NSF CAREER, NSF \#1941722, RAI Institute, ONR grant N00014-22-1-2293, and ONR grant N00014-22-1-2621.

}


\printbibliography

@misc{dong2026expoftsampleefficientreinforcementlearning,
      title={EXPO-FT: Sample-Efficient Reinforcement Learning Finetuning for Vision-Language-Action Models}, 
      author={Perry Dong and Kuo-Han Hung and Tian Gao and Dorsa Sadigh and Chelsea Finn},
      year={2026},
      eprint={2605.25477},
      archivePrefix={arXiv},
      primaryClass={cs.RO},
      url={https://arxiv.org/abs/2605.25477}, 
}

@inproceedings{
    black2025realtime,
    title={Real-Time Execution of Action Chunking Flow Policies},
    author={Kevin Black and Manuel Y Galliker and Sergey Levine},
    booktitle={The Thirty-ninth Annual Conference on Neural Information Processing Systems},
    year={2025},
    url={https://openreview.net/forum?id=UkR2zO5uww}
}

@misc{black2025trainingtimeactionconditioningefficient,
      title={Training-Time Action Conditioning for Efficient Real-Time Chunking}, 
      author={Kevin Black and Allen Z. Ren and Michael Equi and Sergey Levine},
      year={2025},
      eprint={2512.05964},
      archivePrefix={arXiv},
      primaryClass={cs.RO},
      url={https://arxiv.org/abs/2512.05964}, 
}

@misc{
psenka2025learningdiffusionmodelpolicy,
title={Learning a Diffusion Model Policy from Rewards via Q-Score Matching},
author={Michael Psenka and Alejandro Escontrela and Pieter Abbeel and Yi Ma},
year={2024},
url={https://openreview.net/forum?id=StkLULT1i1}
}

@inproceedings{
li2026qlearningadjointmatching,
title={Q-Learning with Adjoint Matching},
author={Qiyang Li and Sergey Levine},
booktitle={The Fourteenth International Conference on Learning Representations},
year={2026},
url={https://openreview.net/forum?id=vd4eNAdtO6}
}

@misc{mark2024policyagnosticrloffline,
      title={Policy Agnostic RL: Offline RL and Online RL Fine-Tuning of Any Class and Backbone}, 
      author={Max Sobol Mark and Tian Gao and Georgia Gabriela Sampaio and Mohan Kumar Srirama and Archit Sharma and Chelsea Finn and Aviral Kumar},
      year={2024},
      eprint={2412.06685},
      archivePrefix={arXiv},
      primaryClass={cs.LG},
      url={https://arxiv.org/abs/2412.06685}, 
}

@misc{dong2026qlearningworldmodels,
      title={Q-Learning With World Models}, 
      author={Perry Dong and Yueru Jia and Chelsea Finn and Dorsa Sadigh},
      year={2026},
      eprint={2608.17163},
      archivePrefix={arXiv},
      primaryClass={cs.LG},
      url={https://arxiv.org/abs/2608.17163}, 
}

@misc{dong2026reallyneedpretrainqfunctions,
      title={Do You Really Need to Pretrain Q-Functions for Online RL Fine-Tuning?}, 
      author={Perry Dong and Ron Polonsky and Dorsa Sadigh and Chelsea Finn},
      year={2026},
      eprint={2607.27203},
      archivePrefix={arXiv},
      primaryClass={cs.LG},
      url={https://arxiv.org/abs/2607.27203}, 
}

@misc{nauman2024biggerregularizedoptimisticscaling,
      title={Bigger, Regularized, Optimistic: scaling for compute and sample-efficient continuous control}, 
      author={Michal Nauman and Mateusz Ostaszewski and Krzysztof Jankowski and Piotr Miłoś and Marek Cygan},
      year={2024},
      eprint={2405.16158},
      archivePrefix={arXiv},
      primaryClass={cs.LG},
      url={https://arxiv.org/abs/2405.16158}, 
}

@inproceedings{
    chen2021randomizedensembleddoubleqlearning,
    title={Randomized Ensembled Double Q-Learning: Learning Fast Without a Model},
    author={Xinyue Chen and Che Wang and Zijian Zhou and Keith W. Ross},
    booktitle={International Conference on Learning Representations},
    year={2021},
    url={https://openreview.net/forum?id=AY8zfZm0tDd}
}

@misc{dong2026tqlscalingqfunctionstransformers,
      title={TQL: Scaling Q-Functions with Transformers by Preventing Attention Collapse}, 
      author={Perry Dong and Kuo-Han Hung and Alexander Swerdlow and Dorsa Sadigh and Chelsea Finn},
      year={2026},
      eprint={2602.01439},
      archivePrefix={arXiv},
      primaryClass={cs.LG},
      url={https://arxiv.org/abs/2602.01439}, 
}

@inproceedings{ball2023efficientonlinereinforcementlearning,
author = {Ball, Philip J. and Smith, Laura and Kostrikov, Ilya and Levine, Sergey},
title = {Efficient online reinforcement learning with offline data},
year = {2023},
publisher = {JMLR.org},
booktitle = {Proceedings of the 40th International Conference on Machine Learning},
articleno = {67},
numpages = {18},
location = {Honolulu, Hawaii, USA},
series = {ICML'23}
}

@inproceedings{
    dong2026expostablereinforcementlearning,
    title={{EXPO}: Stable Reinforcement Learning with Expressive Policies},
    author={Perry Dong and Qiyang Li and Dorsa Sadigh and Chelsea Finn},
    booktitle={The Fourteenth International Conference on Learning Representations},
    year={2026},
    url={https://openreview.net/forum?id=aFjSjkB6CV}
}

@misc{dong2026valueflows,
      title={Value Flows}, 
      author={Perry Dong and Chongyi Zheng and Chelsea Finn and Dorsa Sadigh and Benjamin Eysenbach},
      year={2026},
      eprint={2510.07650},
      archivePrefix={arXiv},
      primaryClass={cs.LG},
      url={https://arxiv.org/abs/2510.07650}, 
}

@misc{chen2026pitextttrlonlinerlfinetuning,
      title={$\pi_\texttt{RL}$: Online RL Fine-tuning for Flow-based Vision-Language-Action Models}, 
      author={Kang Chen and Zhihao Liu and Tonghe Zhang and Zhen Guo and Si Xu and Hao Lin and Hongzhi Zang and Xiang Li and Quanlu Zhang and Zhaofei Yu and Guoliang Fan and Tiejun Huang and Yu Wang and Chao Yu},
      year={2026},
      eprint={2510.25889},
      archivePrefix={arXiv},
      primaryClass={cs.LG},
      url={https://arxiv.org/abs/2510.25889}, 
}

@misc{lu2025vlarlmasterfulgeneralrobotic,
      title={VLA-RL: Towards Masterful and General Robotic Manipulation with Scalable Reinforcement Learning}, 
      author={Guanxing Lu and Wenkai Guo and Chubin Zhang and Yuheng Zhou and Haonan Jiang and Zifeng Gao and Yansong Tang and Ziwei Wang},
      year={2025},
      eprint={2505.18719},
      archivePrefix={arXiv},
      primaryClass={cs.RO},
      url={https://arxiv.org/abs/2505.18719}, 
}

@misc{chen2025conrftreinforcedfinetuningmethod,
      title={ConRFT: A Reinforced Fine-tuning Method for VLA Models via Consistency Policy}, 
      author={Yuhui Chen and Shuai Tian and Shugao Liu and Yingting Zhou and Haoran Li and Dongbin Zhao},
      year={2025},
      eprint={2502.05450},
      archivePrefix={arXiv},
      primaryClass={cs.RO},
      url={https://arxiv.org/abs/2502.05450}, 
}

@misc{ankile2025residualoffpolicyrlfinetuning,
      title={Residual Off-Policy RL for Finetuning Behavior Cloning Policies}, 
      author={Lars Ankile and Zhenyu Jiang and Rocky Duan and Guanya Shi and Pieter Abbeel and Anusha Nagabandi},
      year={2025},
      eprint={2509.19301},
      archivePrefix={arXiv},
      primaryClass={cs.RO},
      url={https://arxiv.org/abs/2509.19301}, 
}

@inproceedings{
xiao2025selfimprovingvisionlanguageactionmodelsdata,
title={Self-Improving Vision-Language-Action Models with Data Generation via Residual {RL}},
author={Wenli Xiao and Haotian Lin and Andy Peng and Haoru Xue and Tairan He and Zhengyi Luo and Yuqi Xie and Fengyuan Hu and Linxi Fan and Guanya Shi and Yuke Zhu},
booktitle={The Fourteenth International Conference on Learning Representations},
year={2026},
url={https://openreview.net/forum?id=eUGoqrZ6Ea}
}

@misc{dong2026fastervalueguidedsamplingfast,
      title={FASTER: Value-Guided Sampling for Fast RL}, 
      author={Perry Dong and Alexander Swerdlow and Dorsa Sadigh and Chelsea Finn},
      year={2026},
      eprint={2604.19730},
      archivePrefix={arXiv},
      primaryClass={cs.LG},
      url={https://arxiv.org/abs/2604.19730}, 
}

@misc{intelligence2025pi06vlalearnsexperience,
      title={$\pi^{*}_{0.6}$: a VLA That Learns From Experience}, 
      author={Physical Intelligence and Ali Amin and Raichelle Aniceto and Ashwin Balakrishna and Kevin Black and Ken Conley and Grace Connors and James Darpinian and Karan Dhabalia and Jared DiCarlo and Danny Driess and Michael Equi and Adnan Esmail and Yunhao Fang and Chelsea Finn and Catherine Glossop and Thomas Godden and Ivan Goryachev and Lachy Groom and Hunter Hancock and Karol Hausman and Gashon Hussein and Brian Ichter and Szymon Jakubczak and Rowan Jen and Tim Jones and Ben Katz and Liyiming Ke and Chandra Kuchi and Marinda Lamb and Devin LeBlanc and Sergey Levine and Adrian Li-Bell and Yao Lu and Vishnu Mano and Mohith Mothukuri and Suraj Nair and Karl Pertsch and Allen Z. Ren and Charvi Sharma and Lucy Xiaoyang Shi and Laura Smith and Jost Tobias Springenberg and Kyle Stachowicz and Will Stoeckle and Alex Swerdlow and James Tanner and Marcel Torne and Quan Vuong and Anna Walling and Haohuan Wang and Blake Williams and Sukwon Yoo and Lili Yu and Ury Zhilinsky and Zhiyuan Zhou},
      year={2025},
      eprint={2511.14759},
      archivePrefix={arXiv},
      primaryClass={cs.LG},
      url={https://arxiv.org/abs/2511.14759}, 
}

@misc{lei2026rl100performantroboticmanipulation,
      title={RL-100: Performant Robotic Manipulation with Real-World Reinforcement Learning}, 
      author={Kun Lei and Huanyu Li and Dongjie Yu and Zhenyu Wei and Lingxiao Guo and Zhennan Jiang and Ziyu Wang and Shiyu Liang and Huazhe Xu},
      year={2026},
      eprint={2510.14830},
      archivePrefix={arXiv},
      primaryClass={cs.RO},
      url={https://arxiv.org/abs/2510.14830}, 
}

@inproceedings{
ren2024diffusionpolicypolicyoptimization,
title={Diffusion Policy Policy Optimization},
author={Allen Z. Ren and Justin Lidard and Lars Lien Ankile and Anthony Simeonov and Pulkit Agrawal and Anirudha Majumdar and Benjamin Burchfiel and Hongkai Dai and Max Simchowitz},
booktitle={The Thirteenth International Conference on Learning Representations},
year={2025},
url={https://openreview.net/forum?id=mEpqHvbD2h}
}

@misc{xu2026rltokenbootstrappingonline,
      title={RL Token: Bootstrapping Online RL with Vision-Language-Action Models}, 
      author={Charles Xu and Jost Tobias Springenberg and Michael Equi and Ali Amin and Adnan Esmail and Sergey Levine and Liyiming Ke},
      year={2026},
      eprint={2604.23073},
      archivePrefix={arXiv},
      primaryClass={cs.LG},
      url={https://arxiv.org/abs/2604.23073}, 
}

@misc{luo2025precisedexterousroboticmanipulation,
      title={Precise and Dexterous Robotic Manipulation via Human-in-the-Loop Reinforcement Learning}, 
      author={Jianlan Luo and Charles Xu and Jeffrey Wu and Sergey Levine},
      year={2025},
      eprint={2410.21845},
      archivePrefix={arXiv},
      primaryClass={cs.RO},
      url={https://arxiv.org/abs/2410.21845}, 
}

@misc{haarnoja2018soft,
      title={Soft Actor-Critic: Off-Policy Maximum Entropy Deep Reinforcement Learning with a Stochastic Actor}, 
      author={Tuomas Haarnoja and Aurick Zhou and Pieter Abbeel and Sergey Levine},
      year={2018},
      eprint={1801.01290},
      archivePrefix={arXiv},
      primaryClass={cs.LG},
      url={https://arxiv.org/abs/1801.01290}, 
}

@misc{luo2025serlsoftwaresuitesampleefficient,
      title={SERL: A Software Suite for Sample-Efficient Robotic Reinforcement Learning}, 
      author={Jianlan Luo and Zheyuan Hu and Charles Xu and You Liang Tan and Jacob Berg and Archit Sharma and Stefan Schaal and Chelsea Finn and Abhishek Gupta and Sergey Levine},
      year={2025},
      eprint={2401.16013},
      archivePrefix={arXiv},
      primaryClass={cs.RO},
      url={https://arxiv.org/abs/2401.16013}, 
}

@misc{geminiroboticsteam2025geminirobotics15pushing,
      title={Gemini Robotics 1.5: Pushing the Frontier of Generalist Robots with Advanced Embodied Reasoning, Thinking, and Motion Transfer}, 
      author={Gemini Robotics Team and Abbas Abdolmaleki and Saminda Abeyruwan and Joshua Ainslie and Jean-Baptiste Alayrac and Montserrat Gonzalez Arenas and Ashwin Balakrishna and Nathan Batchelor and Alex Bewley and Jeff Bingham and Michael Bloesch and Konstantinos Bousmalis and Philemon Brakel and Anthony Brohan and Thomas Buschmann and Arunkumar Byravan and Serkan Cabi and Ken Caluwaerts and Federico Casarini and Christine Chan and Oscar Chang and London Chappellet-Volpini and Jose Enrique Chen and Xi Chen and Hao-Tien Lewis Chiang and Krzysztof Choromanski and Adrian Collister and David B. D'Ambrosio and Sudeep Dasari and Todor Davchev and Meet Kirankumar Dave and Coline Devin and Norman Di Palo and Tianli Ding and Carl Doersch and Adil Dostmohamed and Yilun Du and Debidatta Dwibedi and Sathish Thoppay Egambaram and Michael Elabd and Tom Erez and Xiaolin Fang and Claudio Fantacci and Cody Fong and Erik Frey and Chuyuan Fu and Ruiqi Gao and Marissa Giustina and Keerthana Gopalakrishnan and Laura Graesser and Oliver Groth and Agrim Gupta and Roland Hafner and Steven Hansen and Leonard Hasenclever and Sam Haves and Nicolas Heess and Brandon Hernaez and Alex Hofer and Jasmine Hsu and Lu Huang and Sandy H. Huang and Atil Iscen and Mithun George Jacob and Deepali Jain and Sally Jesmonth and Abhishek Jindal and Ryan Julian and Dmitry Kalashnikov and M. Emre Karagozler and Stefani Karp and Matija Kecman and J. Chase Kew and Donnie Kim and Frank Kim and Junkyung Kim and Thomas Kipf and Sean Kirmani and Ksenia Konyushkova and Li Yang Ku and Yuheng Kuang and Thomas Lampe and Antoine Laurens and Tuan Anh Le and Isabel Leal and Alex X. Lee and Tsang-Wei Edward Lee and Guy Lever and Jacky Liang and Li-Heng Lin and Fangchen Liu and Shangbang Long and Caden Lu and Sharath Maddineni and Anirudha Majumdar and Kevis-Kokitsi Maninis and Andrew Marmon and Sergio Martinez and Assaf Hurwitz Michaely and Niko Milonopoulos and Joss Moore and Robert Moreno and Michael Neunert and Francesco Nori and Joy Ortiz and Kenneth Oslund and Carolina Parada and Emilio Parisotto and Amaris Paryag and Acorn Pooley and Thomas Power and Alessio Quaglino and Haroon Qureshi and Rajkumar Vasudeva Raju and Helen Ran and Dushyant Rao and Kanishka Rao and Isaac Reid and David Rendleman and Krista Reymann and Miguel Rivas and Francesco Romano and Yulia Rubanova and Peter Pastor Sampedro and Pannag R Sanketi and Dhruv Shah and Mohit Sharma and Kathryn Shea and Mohit Shridhar and Charles Shu and Vikas Sindhwani and Sumeet Singh and Radu Soricut and Rachel Sterneck and Ian Storz and Razvan Surdulescu and Jie Tan and Jonathan Tompson and Saran Tunyasuvunakool and Jake Varley and Grace Vesom and Giulia Vezzani and Maria Bauza Villalonga and Oriol Vinyals and René Wagner and Ayzaan Wahid and Stefan Welker and Paul Wohlhart and Chengda Wu and Markus Wulfmeier and Fei Xia and Ted Xiao and Annie Xie and Jinyu Xie and Peng Xu and Sichun Xu and Ying Xu and Zhuo Xu and Jimmy Yan and Sherry Yang and Skye Yang and Yuxiang Yang and Hiu Hong Yu and Wenhao Yu and Wentao Yuan and Yuan Yuan and Jingwei Zhang and Tingnan Zhang and Zhiyuan Zhang and Allan Zhou and Guangyao Zhou and Yuxiang Zhou},
      year={2025},
      eprint={2510.03342},
      archivePrefix={arXiv},
      primaryClass={cs.RO},
      url={https://arxiv.org/abs/2510.03342}, 
}

@misc{dong2025reinforcementlearningimplicitimitation,
      title={Reinforcement Learning via Implicit Imitation Guidance}, 
      author={Perry Dong and Alec M. Lessing and Annie S. Chen and Chelsea Finn},
      year={2025},
      eprint={2506.07505},
      archivePrefix={arXiv},
      primaryClass={cs.LG},
      url={https://arxiv.org/abs/2506.07505}, 
}

@misc{intelligence2025pi05visionlanguageactionmodelopenworld,
      title={$\pi_{0.5}$: a Vision-Language-Action Model with Open-World Generalization}, 
      author={Physical Intelligence and Kevin Black and Noah Brown and James Darpinian and Karan Dhabalia and Danny Driess and Adnan Esmail and Michael Equi and Chelsea Finn and Niccolo Fusai and Manuel Y. Galliker and Dibya Ghosh and Lachy Groom and Karol Hausman and Brian Ichter and Szymon Jakubczak and Tim Jones and Liyiming Ke and Devin LeBlanc and Sergey Levine and Adrian Li-Bell and Mohith Mothukuri and Suraj Nair and Karl Pertsch and Allen Z. Ren and Lucy Xiaoyang Shi and Laura Smith and Jost Tobias Springenberg and Kyle Stachowicz and James Tanner and Quan Vuong and Homer Walke and Anna Walling and Haohuan Wang and Lili Yu and Ury Zhilinsky},
      year={2025},
      eprint={2504.16054},
      archivePrefix={arXiv},
      primaryClass={cs.LG},
      url={https://arxiv.org/abs/2504.16054}, 
}

@inproceedings{
hu2021loralowrankadaptationlarge,
title={Lo{RA}: Low-Rank Adaptation of Large Language Models},
author={Edward J Hu and yelong shen and Phillip Wallis and Zeyuan Allen-Zhu and Yuanzhi Li and Shean Wang and Lu Wang and Weizhu Chen},
booktitle={International Conference on Learning Representations},
year={2022},
url={https://openreview.net/forum?id=nZeVKeeFYf9}
}

@misc{park2026pimathbfr2reactiverealtimeflow,
      title={$\pi\mathbf{R}^2$: Reactive Real-time Flow Policies}, 
      author={Sungjae Park and Shubham Tulsiani},
      year={2026},
      eprint={2607.26055},
      archivePrefix={arXiv},
      primaryClass={cs.RO},
      url={https://arxiv.org/abs/2607.26055}, 
}

@misc{tang2026vlashrealtimevlasfuturestateaware,
      title={VLASH: Real-Time VLAs via Future-State-Aware Asynchronous Inference}, 
      author={Jiaming Tang and Yufei Sun and Yilong Zhao and Shang Yang and Yujun Lin and Zhuoyang Zhang and James Hou and Yao Lu and Zhijian Liu and Song Han},
      year={2026},
      eprint={2512.01031},
      archivePrefix={arXiv},
      primaryClass={cs.RO},
      url={https://arxiv.org/abs/2512.01031}, 
}

@misc{jiang2026futurertcrealtimerobotexecution,
      title={FutureRTC: Real-Time Robot Execution with Anticipatory-Conditioned Action Chunking}, 
      author={Hai Jiang and Yixian Zou and Binbin Liang and Boqian Liu and Fanman Meng and Shuaicheng Liu},
      year={2026},
      eprint={2607.24008},
      archivePrefix={arXiv},
      primaryClass={cs.RO},
      url={https://arxiv.org/abs/2607.24008}, 
}

@misc{sendai2025leaveobservationbehindrealtime,
      title={Leave No Observation Behind: Real-time Correction for VLA Action Chunks}, 
      author={Kohei Sendai and Maxime Alvarez and Tatsuya Matsushima and Yutaka Matsuo and Yusuke Iwasawa},
      year={2025},
      eprint={2509.23224},
      archivePrefix={arXiv},
      primaryClass={cs.RO},
      url={https://arxiv.org/abs/2509.23224}, 
}

@misc{ma2025runningvlasrealtimespeed,
      title={Running VLAs at Real-time Speed}, 
      author={Yunchao Ma and Yizhuang Zhou and Yunhuan Yang and Tiancai Wang and Haoqiang Fan},
      year={2025},
      eprint={2510.26742},
      archivePrefix={arXiv},
      primaryClass={cs.RO},
      url={https://arxiv.org/abs/2510.26742}, 
}

@article{lu2026faster,
  title={FASTER: Rethinking Real-Time Flow VLAs}, 
  author={Yuxiang Lu and Zhe Liu and Xianzhe Fan and Zhenya Yang and Jinghua Hou and Junyi Li and Kaixin Ding and Hengshuang Zhao},
  year={2026},
  journal={arXiv preprint arXiv:2603.19199}
}

@misc{niu2026realtimevlaflashspeculativeinference,
      title={Realtime-VLA FLASH: Speculative Inference Framework for Diffusion-based VLAs}, 
      author={Jiahui Niu and Kefan Gu and Yucheng Zhao and Shengwen Liang and Tiancai Wang and Xing Hu and Ying Wang and Huawei Li},
      year={2026},
      eprint={2605.13778},
      archivePrefix={arXiv},
      primaryClass={cs.RO},
      url={https://arxiv.org/abs/2605.13778}, 
}

@inproceedings{
yang2025efficientvla,
title={Efficient{VLA}: Training-Free Acceleration and Compression for Vision-Language-Action Models},
author={Yantai Yang and Yuhao Wang and Zichen Wen and Luo Zhongwei and Chang Zou and Zhipeng Zhang and Chuan Wen and Linfeng Zhang},
booktitle={The Thirty-ninth Annual Conference on Neural Information Processing Systems},
year={2025},
url={https://openreview.net/forum?id=SELYlDHZk2}
}

@misc{nvidia2025gr00tn1openfoundation,
      title={GR00T N1: An Open Foundation Model for Generalist Humanoid Robots}, 
      author={NVIDIA and : and Johan Bjorck and Fernando Castañeda and Nikita Cherniadev and Xingye Da and Runyu Ding and Linxi "Jim" Fan and Yu Fang and Dieter Fox and Fengyuan Hu and Spencer Huang and Joel Jang and Zhenyu Jiang and Jan Kautz and Kaushil Kundalia and Lawrence Lao and Zhiqi Li and Zongyu Lin and Kevin Lin and Guilin Liu and Edith Llontop and Loic Magne and Ajay Mandlekar and Avnish Narayan and Soroush Nasiriany and Scott Reed and You Liang Tan and Guanzhi Wang and Zu Wang and Jing Wang and Qi Wang and Jiannan Xiang and Yuqi Xie and Yinzhen Xu and Zhenjia Xu and Seonghyeon Ye and Zhiding Yu and Ao Zhang and Hao Zhang and Yizhou Zhao and Ruijie Zheng and Yuke Zhu},
      year={2025},
      eprint={2503.14734},
      archivePrefix={arXiv},
      primaryClass={cs.RO},
      url={https://arxiv.org/abs/2503.14734}, 
}

@misc{patil2026ogposampleefficientfullfinetuning,
      title={OGPO: Sample Efficient Full-Finetuning of Generative Control Policies}, 
      author={Sarvesh Patil and Mitsuhiko Nakamoto and Manan Agarwal and Shashwat Saxena and Jesse Zhang and Giri Anantharaman and Cleah Winston and Chaoyi Pan and Douglas Chen and Nai-Chieh Huang and Zeynep Temel and Oliver Kroemer and Sergey Levine and Abhishek Gupta and Hongkai Dai and Paarth Shah and Max Simchowitz},
      year={2026},
      eprint={2605.03065},
      archivePrefix={arXiv},
      primaryClass={cs.LG},
      url={https://arxiv.org/abs/2605.03065}, 
}

@misc{zhang2026reinflowfinetuningflowmatching,
      title={ReinFlow: Fine-tuning Flow Matching Policy with Online Reinforcement Learning}, 
      author={Tonghe Zhang and Chao Yu and Sichang Su and Yu Wang},
      year={2026},
      eprint={2505.22094},
      archivePrefix={arXiv},
      primaryClass={cs.RO},
      url={https://arxiv.org/abs/2505.22094}, 
}

@INPROCEEDINGS{drloffpolicy,
  author={Gu, Shixiang and Holly, Ethan and Lillicrap, Timothy and Levine, Sergey},
  booktitle={2017 IEEE International Conference on Robotics and Automation (ICRA)}, 
  title={Deep reinforcement learning for robotic manipulation with asynchronous off-policy updates}, 
  year={2017},
  volume={},
  number={},
  pages={3389-3396},
  doi={10.1109/ICRA.2017.7989385}
}

@misc{black2026pi0visionlanguageactionflowmodel,
      title={$\pi_0$: A Vision-Language-Action Flow Model for General Robot Control}, 
      author={Kevin Black and Noah Brown and Danny Driess and Adnan Esmail and Michael Equi and Chelsea Finn and Niccolo Fusai and Lachy Groom and Karol Hausman and Brian Ichter and Szymon Jakubczak and Tim Jones and Liyiming Ke and Sergey Levine and Adrian Li-Bell and Mohith Mothukuri and Suraj Nair and Karl Pertsch and Lucy Xiaoyang Shi and James Tanner and Quan Vuong and Anna Walling and Haohuan Wang and Ury Zhilinsky},
      year={2026},
      eprint={2410.24164},
      archivePrefix={arXiv},
      primaryClass={cs.LG},
      url={https://arxiv.org/abs/2410.24164}, 
}

@inproceedings{
matthews2025kinetix,
title={Kinetix: Investigating the Training of General Agents through Open-Ended Physics-Based Control Tasks},
author={Michael Matthews and Michael Beukman and Chris Lu and Jakob Nicolaus Foerster},
booktitle={The Thirteenth International Conference on Learning Representations},
year={2025},
url={https://openreview.net/forum?id=zCxGCdzreM}
}

@misc{khazatsky2025droidlargescaleinthewildrobot,
      title={DROID: A Large-Scale In-The-Wild Robot Manipulation Dataset}, 
      author={Alexander Khazatsky and Karl Pertsch and Suraj Nair and Ashwin Balakrishna and Sudeep Dasari and Siddharth Karamcheti and Soroush Nasiriany and Mohan Kumar Srirama and Lawrence Yunliang Chen and Kirsty Ellis and Peter David Fagan and Joey Hejna and Masha Itkina and Marion Lepert and Yecheng Jason Ma and Patrick Tree Miller and Jimmy Wu and Suneel Belkhale and Shivin Dass and Huy Ha and Arhan Jain and Abraham Lee and Youngwoon Lee and Marius Memmel and Sungjae Park and Ilija Radosavovic and Kaiyuan Wang and Albert Zhan and Kevin Black and Cheng Chi and Kyle Beltran Hatch and Shan Lin and Jingpei Lu and Jean Mercat and Abdul Rehman and Pannag R Sanketi and Archit Sharma and Cody Simpson and Quan Vuong and Homer Rich Walke and Blake Wulfe and Ted Xiao and Jonathan Heewon Yang and Arefeh Yavary and Tony Z. Zhao and Christopher Agia and Rohan Baijal and Mateo Guaman Castro and Daphne Chen and Qiuyu Chen and Trinity Chung and Jaimyn Drake and Ethan Paul Foster and Jensen Gao and Vitor Guizilini and David Antonio Herrera and Minho Heo and Kyle Hsu and Jiaheng Hu and Muhammad Zubair Irshad and Donovon Jackson and Charlotte Le and Yunshuang Li and Kevin Lin and Roy Lin and Zehan Ma and Abhiram Maddukuri and Suvir Mirchandani and Daniel Morton and Tony Nguyen and Abigail O'Neill and Rosario Scalise and Derick Seale and Victor Son and Stephen Tian and Emi Tran and Andrew E. Wang and Yilin Wu and Annie Xie and Jingyun Yang and Patrick Yin and Yunchu Zhang and Osbert Bastani and Glen Berseth and Jeannette Bohg and Ken Goldberg and Abhinav Gupta and Abhishek Gupta and Dinesh Jayaraman and Joseph J Lim and Jitendra Malik and Roberto Martín-Martín and Subramanian Ramamoorthy and Dorsa Sadigh and Shuran Song and Jiajun Wu and Michael C. Yip and Yuke Zhu and Thomas Kollar and Sergey Levine and Chelsea Finn},
      year={2025},
      eprint={2403.12945},
      archivePrefix={arXiv},
      primaryClass={cs.RO},
      url={https://arxiv.org/abs/2403.12945}, 
}

@article{wagenmaker2025steering,
  author    = {Wagenmaker, Andrew and Nakamoto, Mitsuhiko and Zhang, Yunchu and Park, Seohong and Yagoub, Waleed and Nagabandi, Anusha and Gupta, Abhishek and Levine, Sergey},
  title     = {Steering Your Diffusion Policy with Latent Space Reinforcement Learning},
  journal   = {Conference on Robot Learning},
  year      = {2025},
}

@misc{levine2016endtoendtrainingdeepvisuomotor,
      title={End-to-End Training of Deep Visuomotor Policies}, 
      author={Sergey Levine and Chelsea Finn and Trevor Darrell and Pieter Abbeel},
      year={2016},
      eprint={1504.00702},
      archivePrefix={arXiv},
      primaryClass={cs.LG},
      url={https://arxiv.org/abs/1504.00702}, 
}

@misc{mandlekar2020irisimplicitreinforcementinteraction,
      title={IRIS: Implicit Reinforcement without Interaction at Scale for Learning Control from Offline Robot Manipulation Data}, 
      author={Ajay Mandlekar and Fabio Ramos and Byron Boots and Silvio Savarese and Li Fei-Fei and Animesh Garg and Dieter Fox},
      year={2020},
      eprint={1911.05321},
      archivePrefix={arXiv},
      primaryClass={cs.RO},
      url={https://arxiv.org/abs/1911.05321}, 
}

@misc{haarnoja2019softactorcriticalgorithmsapplications,
      title={Soft Actor-Critic Algorithms and Applications}, 
      author={Tuomas Haarnoja and Aurick Zhou and Kristian Hartikainen and George Tucker and Sehoon Ha and Jie Tan and Vikash Kumar and Henry Zhu and Abhishek Gupta and Pieter Abbeel and Sergey Levine},
      year={2019},
      eprint={1812.05905},
      archivePrefix={arXiv},
      primaryClass={cs.LG},
      url={https://arxiv.org/abs/1812.05905}, 
}

@misc{zhu2018dexterousmanipulationdeepreinforcement,
      title={Dexterous Manipulation with Deep Reinforcement Learning: Efficient, General, and Low-Cost}, 
      author={Henry Zhu and Abhishek Gupta and Aravind Rajeswaran and Sergey Levine and Vikash Kumar},
      year={2018},
      eprint={1810.06045},
      archivePrefix={arXiv},
      primaryClass={cs.AI},
      url={https://arxiv.org/abs/1810.06045}, 
}

@misc{sharma2023selfimprovingrobotsendtoendautonomous,
      title={Self-Improving Robots: End-to-End Autonomous Visuomotor Reinforcement Learning}, 
      author={Archit Sharma and Ahmed M. Ahmed and Rehaan Ahmad and Chelsea Finn},
      year={2023},
      eprint={2303.01488},
      archivePrefix={arXiv},
      primaryClass={cs.RO},
      url={https://arxiv.org/abs/2303.01488}, 
}

\arxiv{
\clearpage
\section*{Appendix}

\begin{figure*}[t]
\centering
\includegraphics[width=\textwidth]{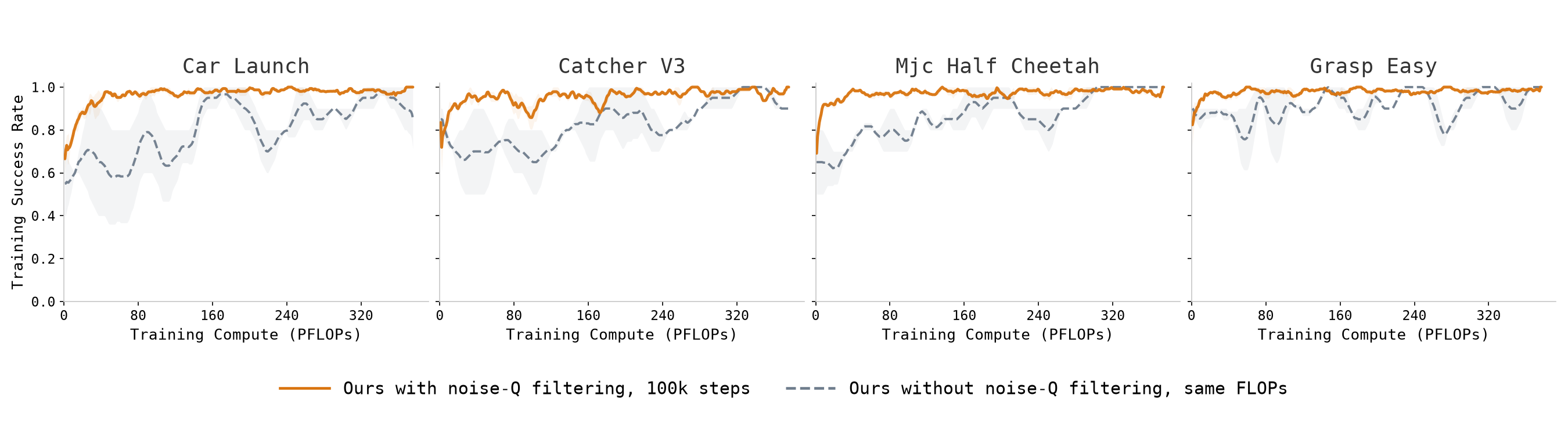}
\caption{\textbf{Comparison of noise-level filtering.} We compare \methodname with and without noise-level filtering, where the latter uses only Q-level filtering during the Bellman backup. The x-axis shows training compute, allowing us to compare the efficiency of the two approaches.}
\label{fig:noise_ablation}
\end{figure*}

\subsection{Noise-level Filtering Studies}
\label{sec:app_value}
To handle the high randomness of dynamic tasks, we empirically find that a larger sampling number such as $N=32$ is often required. However, setting a large $N$ such as 32 introduces a substantial computational burden during training. To address this issue, we incorporate the noise-level filtering technique introduced in \Cref{eq:noise} to filter action candidates during Bellman updates. 

To evaluate the effectiveness and efficiency of noise-level filtering, we compare using and not using noise-level filtering and instead filters over the fully denoised actions. As shown in the \Cref{fig:noise_ablation}, using noise filtering learns significantly more efficiently than not using it under the same training compute, demonstrating that noise-level filtering can effectively improve training by enabling a large $N$ without incurring the computational cost.

\subsection{Full Simulation Experiment Results}
\label{sec:app_sim_results}

Here, we provide detailed simulation experiment results, including evaluations of all baselines. For each method, we conduct 100 trials in each environment and average the results across four random seeds. As shown in \Cref{tab:full_sim_results}, \methodname outperforms all baselines across the evaluated environments.

\begin{figure*}[t]
\centering
\includegraphics[width=\textwidth]{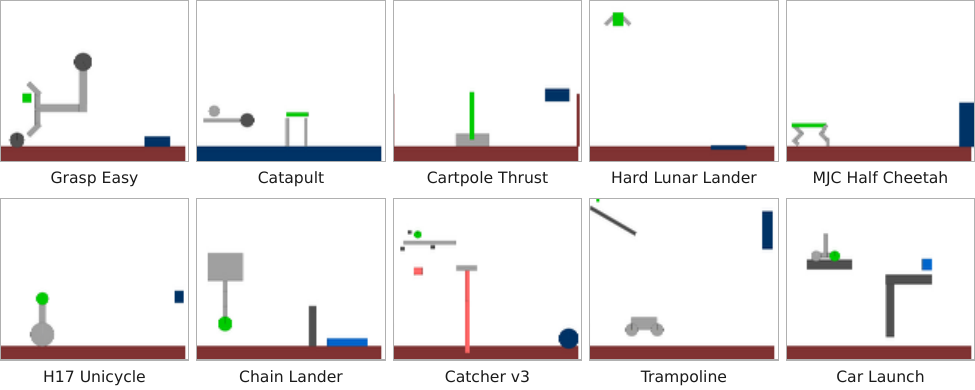}
\caption{\textbf{10 Kinetix simulation tasks \cite{matthews2025kinetix} evaluated in our experiments.}}
\label{fig:sim_tasks}
\end{figure*}

\begin{table*}[t]
\centering
\setlength{\tabcolsep}{4pt}
\caption{Full simulation results on the Kinetix benchmark: success rate (\%). RL results average four random seeds $\times$ 100 evaluation episodes; BC deploys the pretrained policy without fine-tuning, and RTC adds real-time chunking on top of it (512 episodes each). Delay-4 methods replan every 4 steps under a 4-step inference delay; RLPD supports only zero delay. \textbf{Bold} marks every RL method within $0.95\times$ the best RL result on that task (BC and RTC are excluded from the comparison).}
\label{tab:full_sim_results}
\begin{tabular*}{\textwidth}{@{\extracolsep{\fill}}lccccccccc}
\toprule
\multirow{2}{*}{Task} & \multicolumn{2}{c}{Delay $=0$} & \multicolumn{7}{c}{Delay $=4$} \\
\cmidrule(lr){2-3}\cmidrule(lr){4-10}
& BC & RLPD & BC & RTC & DSRL & DSRL w/ RTC & EXPO-FT & EXPO-FT w/ RTC & \methodname \\
\midrule
Car Launch         & 94\%  & \textbf{97\%} & 51\% & 70\% & 76\% & 79\% & 44\% & 64\% & \textbf{98\%} \\
Cartpole Thrust    & 100\% & 57\% & 37\% & 92\% & 49\% & 85\% & \textbf{99\%} & \textbf{96\%} & \textbf{98\%} \\
Catapult           & 62\%  & 83\% & 31\% & 42\% & 36\% & 49\% & 18\% & 47\% & \textbf{93\%} \\
Catcher            & 97\%  & 90\% & 23\% & 90\% & 40\% & 71\% & 90\% & 85\% & \textbf{97\%} \\
Unicycle           & 97\%  & 90\% & 58\% & 74\% & 48\% & 86\% & 48\% & 87\% & \textbf{99\%} \\
Hard Lunar Lander  & 95\%  & 70\% & 57\% & 87\% & 75\% & 79\% & \textbf{95\%} & 90\% & \textbf{94\%} \\
Half-Cheetah       & 88\%  & 93\% & 72\% & 75\% & 86\% & 85\% & 75\% & 79\% & \textbf{98\%} \\
Trampoline         & 84\%  & 47\% & 68\% & 85\% & 43\% & 42\% & \textbf{93\%} & 83\% & \textbf{94\%} \\
Chain Lander       & 95\%  & 90\% & 94\% & 94\% & \textbf{92\%} & \textbf{96\%} & 90\% & \textbf{94\%} & \textbf{92\%} \\
Grasp              & 95\%  & \textbf{97\%} & 61\% & 91\% & 72\% & 89\% & \textbf{97\%} & 92\% & \textbf{99\%} \\
\midrule
Average            & 90.7\% & 81.4\% & 55.2\% & 80.0\% & 61.7\% & 76.1\% & 74.9\% & 81.7\% & \textbf{96.2\%} \\
\bottomrule
\end{tabular*}
\end{table*}


\subsection{Detailed Simulation Task Settings}
\label{appendix:sim_settings}

\subsubsection{Environment and Base Policy}
\label{sec:app_sim_environment}

For simulation, we use the vector-state (symbolic) Kinetix benchmark, where no VLA is involved and the base policy is a pretrained state-based flow-matching policy. We evaluate on 10 environments:
\texttt{car\_launch}, \texttt{cartpole\_thrust},
\texttt{catapult}, \texttt{catcher\_v3}, \texttt{h17\_unicycle}, \texttt{hard\_lunar\_lander},
\texttt{mjc\_half\_cheetah}, \texttt{trampoline}, \texttt{chain\_lander}, and
\texttt{grasp\_easy}.
We use four random seeds and a budget of 100k environment steps per run. The environment applies Gaussian action noise with a standard deviation of 0.1, matching the reference data generation and evaluation rollouts. Success is determined by the environment's episode-solved flag.

The base policy is the publicly released per-level behavior-cloned flow policy from the real-time-chunking \cite{black2025realtime} Kinetix benchmark \cite{matthews2025kinetix}. It uses a channel dimension of 256, a channel hidden dimension of 512, a token hidden dimension of 64, four layers, an action-chunk length of $H=8$, and five flow-matching steps during training. The policy is not delay-conditioned. At rollout and during the critic backup, we sample the policy using 10 Euler denoising steps.

Online fine-tuning of the base policy uses AdamW with a learning rate of $3\times10^{-4}$, weight decay of $10^{-2}$, gradient-norm clipping of 10, and a 1000-step warmup. We use a prefix-conditioned flow-matching objective with
$ d \sim \mathrm{Unif}\{0,\ldots,4\} $, where the prefix length is resampled independently for each training example. Unlike the real-world setting, where the online prefix length is fixed to the deployment delay, the simulation setting resamples the prefix length during training.

\subsubsection{Learner Configuration}
\label{sec:app_sim_learner}

The critic, filter, and edit policy use the same overall architecture as in the real-world setting, including a REDQ ensemble of 10 networks with two networks subsampled for each target estimate, LayerNorm, and hidden dimensions $(256,256,256)$. For simulation, the visual encoder is replaced by an MLP state encoder with a 256-dimensional output. Different from the real-world settings, the base-policy update is not restricted to successful episodes, the BC loss \Cref{eq:bc} is applied to all the rollout and demo data.

\subsubsection{State-Based Baselines}
\label{sec:app_sim_baselines}

\paragraph{DSRL (state)}
DSRL applies SAC in the flow policy's noise space over the frozen per-level base policy, with $M=1.5$, target entropy 0, no entropy term in the Bellman backup, and no demonstration data. We evaluate three settings: no inference delay; delay $d=4$ with real-time chunking, where the in-flight prefix is inpainted into chunk positions $[0:4]$ and the window $[4:8]$ is executed; and delay $d=4$ with naive replanning, which uses the same stale observation without prefix conditioning. The third setting isolates the effect of prefix conditioning from the effect of acting on stale observations.

\paragraph{RLPD (state)}
RLPD uses SAC and learns directly from scratch, with 50\% demonstration data in every batch and zero inference delay. The pretrained flow checkpoint is used only for observation preprocessing and reference evaluation and does not contribute to the learned policy. Critic hyperparameters are identical to those of \methodname, so the two methods differ only in their actor parameterization and use of the pretrained policy.

\subsection{Detailed Real-World Task Settings}
\label{appendix:task_setting}

\subsubsection{Task Setting Description}
\label{sec:app_task_description}

Here, we provide detailed task settings for the four real-world tasks evaluated in our experiments, including the task objectives, success detector implementation and definition, reward function, initial-state randomization, camera configuration, and demonstration collection procedure.

All four real-world tasks use the same single-arm DROID~\citep{khazatsky2025droidlargescaleinthewildrobot} setup with a $30$\,Hz control rate and two policy camera views, consisting of one exterior camera and one wrist-mounted camera. Each image is resized to $224\times224$. Rewards are sparse and binary: an automatic detector emits $r=1$ on the step at which it declares success and terminates the episode. A timeout terminates the episode with $r=0$ and is treated as a failure. Episodes are additionally capped at a task-specific horizon.

\subsubsection{Success Detectors}
\label{sec:app_success_detectors}

All success detectors operate on the robot's own observation stream and therefore require no external instrumentation. Dynamic Picking is detected proprioceptively: a successful lift is declared when the end-effector height exceeds $0.30$ while the gripper is closed beyond a mid-aperture threshold for 5 consecutive steps. Soccer Kicking and Ball Balancing use analogous wrist-view-based detectors, with Ball Balancing additionally requiring the ball to remain within a specified center tolerance for 10 consecutive frames. Object Passing uses a wrist-view-based detector to determine whether the robot has successfully grasped the object: a success is declared when the gripper is closed and the object remains detected for 5 consecutive control steps.

\subsubsection{Additional Critic Inputs}
\label{sec:app_privileged_inputs}

For two tasks, a small number of quantities already measured by the success detector are written into unused slots of the proprioceptive state vector. These quantities are available to the critic, the noise-$Q$ filter, and the edit policy. The base VLA's own state input remains unchanged, so the supervised checkpoint and normalization statistics are unaffected.

Ball Balancing exposes the plate center, ball position, and ball velocity. Soccer Kicking exposes the keeper's position and velocity. 

For Ball Balancing, we additionally remove the vertical ($z$) proprioceptive dimension from the critic input because it drifts monotonically with episode time and may allow the value function to exploit episode-time information. Both DSRL and RLPD receive the same privileged state dimensions in their critics.

\subsubsection{Observation Layout}
\label{sec:app_observation_layout}

Three tasks use one exterior view and one wrist view. The critic encoder consumes these views as a six-channel tensor. Ball Balancing instead uses a three-frame stack of the exterior view at $t$, $t-k$, and $t-2k$, while dropping the wrist view from the policy input. Consequently, the critic encoder consumes nine channels, and the VLA receives the corresponding three-image configuration.

\subsubsection{Task Full Execution Strips}
\label{sec:task_strips}

We visualize the full execution trajectories of the evaluated tasks. As shown in \Cref{fig:task_strips}, each strip illustrates the temporal progression of the task from initiation to completion, providing a qualitative view of the robot's behavior throughout the entire execution.

\begin{figure*}[t]
\centering
\includegraphics[width=\textwidth]{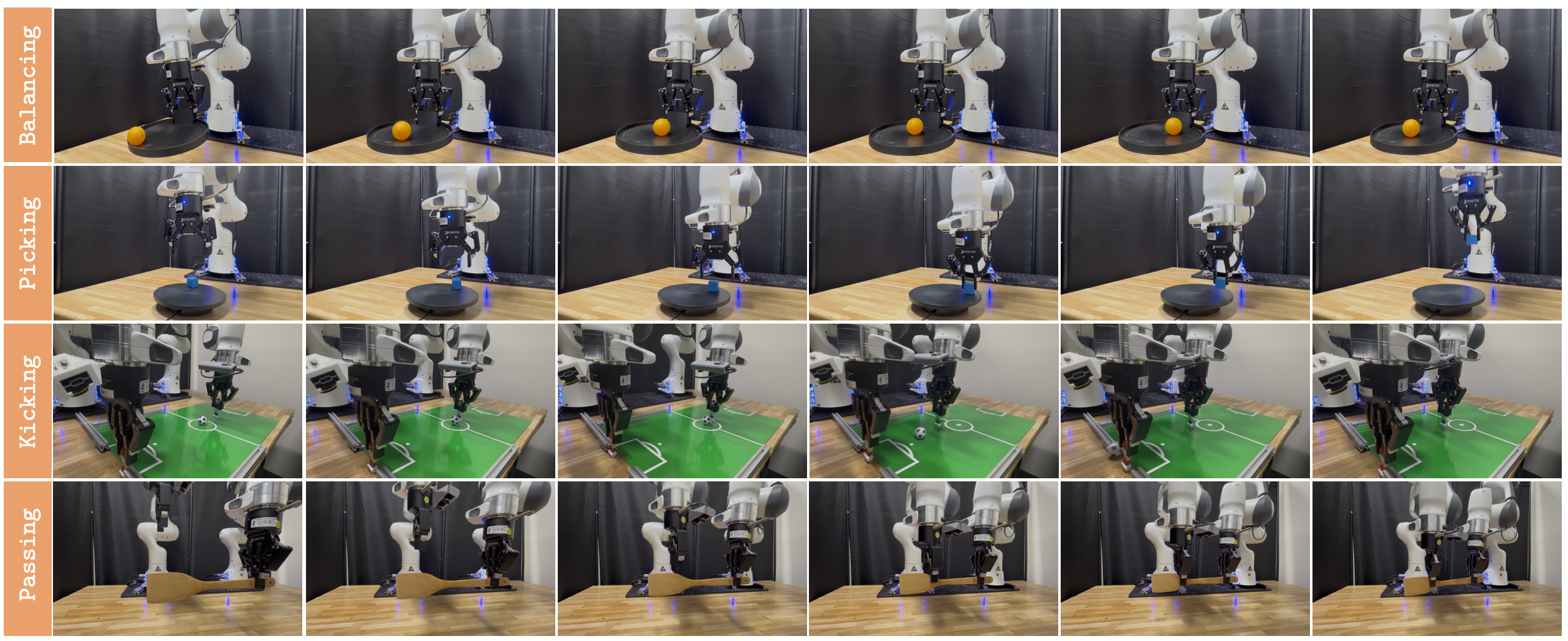}
\caption{Full execution strips for the evaluated tasks. Each strip shows the temporal progression of a complete task execution, from initialization to successful completion.}
\label{fig:task_strips}
\end{figure*}

\subsubsection{Task Initial-State Randomization Space}
\label{sec:app_randomization}

We visualize the task initial-state randomization space to illustrate the range of initial object positions. As shown in \Cref{fig:task_randomization}, the randomized space is highlighted by the orange boxes, capturing the randomization applied to both the objects and the robot.

\begin{figure*}[t]
\centering
\includegraphics[width=\textwidth]{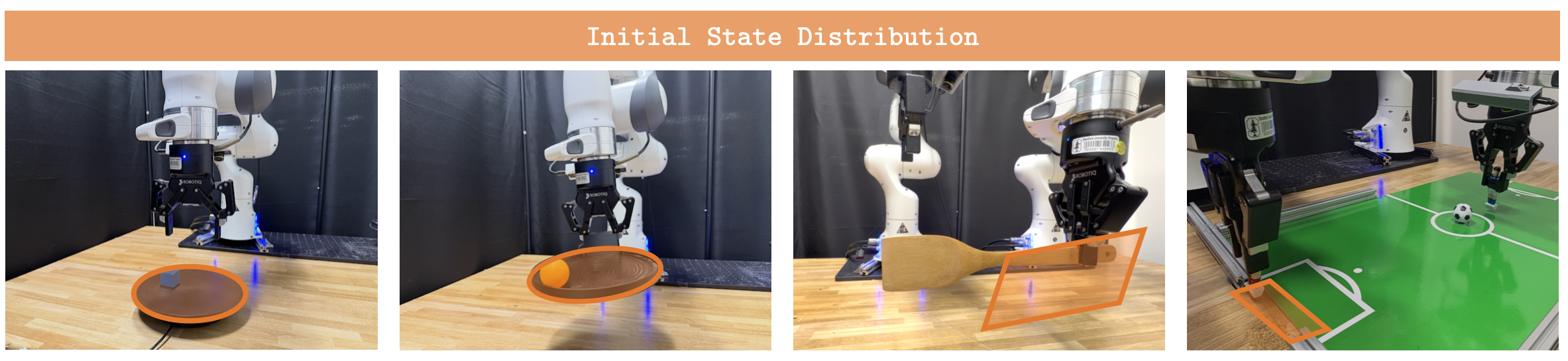}
\caption{Initial-state randomization spaces for the real-world tasks. The orange boxes indicate the regions within which object and robot initial states are randomized.}
\label{fig:task_randomization}
\end{figure*}

\begin{table*}[t]
\centering
\caption{Shared optimization hyperparameters used across all real-world experiments.}
\label{tab:realworld_shared_hyperparams}
\begin{tabular*}{\textwidth}{@{\extracolsep{\fill}}lc}
\toprule
Hyperparameter & Value \\
\midrule
Optimizer (critic, filter, edit policy, temperature)
& Adam, $3\times10^{-4}$ \\
Optimizer (base VLA)
& AdamW, $2.5\times10^{-5}$, clip $1.0$ \\
Critic target update $\tau_Q$
& $5\times10^{-3}$ \\
Base-policy Polyak copy $\tau_\pi$
& $10^{-3}$ \\
Initial temperature $\alpha_0$
& $0.01$ \\
Target entropy
& $-D/2$, $D=C\times7$ \\
Critic minibatch size
& $64$ \\
Update-to-data ratio
& $20$ \\
$Q$-ensemble size / subsample
& $10/2$ \\
Filter-critic ensemble size
& $2$ \\
Action-chunk horizon $H$
& $16$ \\
Base candidates $N$ / edit candidates
& $32/32$ \\
Backup noise seeds / survivors / edits
& $32/1/1$ \\
Denoising steps
& $10$ \\
Image / state embedding dimensions
& $512/64$ \\
Hidden layers
& $(256,256,256)$ \\
\bottomrule
\end{tabular*}
\end{table*}

\begin{table*}[t]
\centering
\caption{Task-specific hyperparameters for \methodname. $K$ denotes the number of collected transitions per update call. ``Prior data'' indicates whether demonstrations are sampled as a fixed fraction of each critic batch or seeded into the online replay buffer. Environment steps denote the total budget of the reported run.}
\label{tab:realworld_task_hyperparams}
\begin{tabular*}{\textwidth}{@{\extracolsep{\fill}}lcccccc}
\toprule
Task & Edit scale & Replan $C$ & Delay $d$ & $K$ & Prior data & Env. steps \\
\midrule
Dynamic Picking
& $0.1$ & $8$ & $3$ & $25$ & Seeded in buffer & $\sim18$k \\
Soccer Kicking
& $0.05$ & $8$ & $5$ & $20$ & $50\%$ of each batch & $\sim18$k \\
Ball Balancing
& $0.1$ & $8$ & $5$ & $30$ & Seeded in buffer & $\sim18$k \\
Object Passing
& $0.1$ & $8$ & $5$ & $30$ & Seeded in buffer & $\sim5$k \\
\bottomrule
\end{tabular*}
\end{table*}

\subsection{Detailed Training Settings}
\label{sec:app_training}

\subsubsection{Base Policy Initialization}
\label{sec:app_base_initialization}

We instantiate \methodname with $\pi_{0.5}$~\citep{intelligence2025pi05visionlanguageactionmodelopenworld} as the base policy. The model uses a LoRA~\citep{hu2021loralowrankadaptationlarge} configuration with a \texttt{gemma\_2b\_lora} language backbone and a \texttt{gemma\_300m\_lora} action expert. The padded action dimension is 32, the action horizon is $H=16$, and the output action dimension is 7. Proprioceptive state is provided to the VLA in Cartesian form, and input images are resized to $224\times224$.

The initialization is a task-specific prefix-conditioned real-time-chunking LoRA supervised fine-tuning of $\pi_{0.5}$ using the corresponding task demonstrations and normalization statistics. During supervised training, the per-example prefix length is sampled as $d \sim \mathrm{Unif}\{0,\ldots,d_{\max}\}.$
The first $d$ chunk positions are provided with clean ground-truth actions at flow time $\tau=1$, and the flow-matching loss is applied only to the remaining $H-d$ positions. This exposes the model to both the boot regime ($d=0$) and the delayed inpainting regime during supervised training. The image encoder is trainable during this stage.

\subsubsection{Model Structure and Learned Components}
\label{sec:app_model}

\paragraph{Value function}
The critic is a REDQ-style~\citep{chen2021randomizedensembleddoubleqlearning} ensemble of 10 $Q$-networks with LayerNorm and three hidden layers of width 256. Every $Q$ evaluation, including both the Bellman target and rollout-time action selection, samples two networks uniformly from the target ensemble and takes their minimum. The target ensemble follows the online ensemble using Polyak averaging with $\tau_Q=5\times10^{-3}$. For Kick and Balance, we additionally train the critic on reward windows containing terminal transitions, whose targets consist only of the observed reward.

\paragraph{Critic visual encoder}
The critic uses a pre-activation ResNetV2 with basic, non-bottleneck residual blocks, stage depths $(3,4,6,3)$, GroupNorm with four groups, and 64 base filters that double at each stage to 512. At $224\times224$ resolution, the stem consists of a stride-2 $7\times7$ convolution followed by max pooling. A single encoder consumes the camera views as a channel-stacked tensor, using six channels for two views and nine channels for Balance's three-frame stack. The resulting representation is projected to a 512-dimensional image embedding using a Dense+LayerNorm head. Proprioception is embedded into 64 dimensions and concatenated with the image embedding and flattened action chunk for $Q$-value prediction. 

\paragraph{Edit policy}
The edit policy is a tanh-squashed Gaussian over the flattened execution window, with dimension $D=C\times7$. For $C=8$, this gives $D=56$. The policy is conditioned on the critic's image embedding, proprioceptive embedding, and the base action chunk being corrected. It reuses the critic's image encoder and contains three hidden layers of width 256. Its output lies in $[-1,1]^D$ and is multiplied by a task-specific edit scale before being added to the base action chunk. For Dynamic Pick, the rotational components of the edit are masked to zero, so the edit acts only on translation and gripper dimensions.

\paragraph{Action selection}
At each replan boundary, the base policy draws $N=32$ stochastic action chunks using 10 Euler denoising steps. Because the VLM prefix is shared across noise samples, it is computed only once. Each base chunk receives one sampled edit, producing 64 candidates in total: 32 base candidates and 32 edited candidates. The executed chunk is selected deterministically using the $\arg\max$ of the minimum-over-two-subsampled target $Q$ value. No softmax is applied over candidates.

\paragraph{Noise-$Q$ backup filter}
Denoising 32 candidates during every Bellman backup would substantially increase computational cost. We therefore pre-filter candidates in noise space. A filter critic $Q_f(s',\epsilon)$ scores 32 raw Gaussian seeds in the padded model action space ($H\times32$). The highest-scoring seed is denoised using $\arg\max$ with a sampling temperature of zero, and one edit candidate is sampled from the resulting action chunk. The outer target-$Q$ maximization then considers one base candidate and one edited candidate.

$Q_f$ is a two-network ensemble trained at every critic step using MSE regression onto the outer target critic's $Q$ value for the denoised survivor, with the target stop-gradient applied. Because the regression target is supervised, $Q_f$ does not require a target network. The filter additionally conditions on the delayed observation used by the base policy, consisting of the image embedding and proprioception. Rollout action selection is unaffected by this filter.

\paragraph{Base-policy fine-tuning}
The base VLA is fine-tuned using prefix-conditioned flow-matching behavior cloning on successful episodes, including demonstrations and successful online episodes. Exactly one base-policy update is performed per update call. The prefix length is deterministic during online training: $d=0$ for transitions in the first chunk of an episode and $d$ equal to the deployment delay thereafter. Unlike supervised initialization, the online prefix length is therefore not resampled.

The trainable parameters include the LoRA adapters in the language model as well as all parameters outside the frozen non-LoRA language-model weights, including the SigLIP vision tower and projection layers. The frozen language-model parameters are maintained in bfloat16. The VLA is optimized using the AdamW configuration of the underlying implementation~\citep{intelligence2025pi05visionlanguageactionmodelopenworld}, with $\beta_1=0.9$, $\beta_2=0.95$, $\epsilon=10^{-8}$, weight decay $10^{-10}$, and gradient-norm clipping at 1.0. We use a constant learning rate of $2.5\times10^{-5}$ and no EMA. A Polyak copy of the base-policy parameters is maintained with $\tau_\pi=10^{-3}$ but is not used by the current Bellman backup, which samples next actions from the live base policy.

\paragraph{Entropy and temperature}
The learnable temperature is initialized at $\alpha_0=0.01$ and optimized with Adam at $3\times10^{-4}$ using a target entropy of $-D/2$, where $D=C\times7$ is the edit-policy dimension. Entropy affects only the edit-policy objective and does not appear in the Bellman backup.

\paragraph{Image augmentation}
Both current and next observations are augmented independently. For each view, we apply a 95\% random crop followed by resizing to $224\times224$, a random rotation in $[-5^\circ,5^\circ]$, and color jitter with brightness, contrast, and saturation changes of $\pm0.1$. The same augmentation procedure is applied to critic, filter, edit-policy, and base-policy inputs.

\subsubsection{Latency Model}
\label{sec:app_latency}

We study two ways of realizing inference latency at a $30$ Hz control rate. In the \emph{wall-clock} condition, we set $d=0$ and add $100$ ms of real sleep to every sample actions call for soccer kicking, ball balancing, and object passing, approximating the compute time of running the $\pi_{0.5}$ model on a typical edge GPU. For dynamic picking, we do not inject additional wall-clock latency, as the task is highly dynamic and even modest additional latency causes the baseline methods to fail almost entirely, making the comparison less informative.  In the \emph{chunk-delay} condition, the policy observes a $d$-step-old observation, the $d$ actions currently in flight are inpainted into chunk positions $[0,d)$ as a clean prefix, and the window $[d,d+C)$ is executed. We use $d=3$ for dynamic picking and $d=5$ for the other tasks, corresponding to approximately $100$ ms and $167$ ms, respectively, and matching the inference-time settings used in the wall-clock condition.

\subsubsection{Optimization Hyperparameters}
\label{sec:app_optimization}

Table~\ref{tab:realworld_shared_hyperparams} lists the hyperparameters shared across all real-world experiments, while Table~\ref{tab:realworld_task_hyperparams} lists the task-specific settings.

All learned components other than the base VLA, including the $Q$ ensemble, filter critic, edit policy, and temperature, use Adam with a learning rate of $3\times10^{-4}$. The base VLA uses the AdamW configuration described above. Each update call samples $\text{batch size}\times\text{UTD}$ transitions and performs the specified number of critic gradient steps on disjoint minibatches, followed by exactly one base-policy step, one edit-policy step, and one temperature step. Thus, the base-policy-to-critic gradient-step ratio is $1:20$ rather than $1:1$.

Update calls are accumulated at a rate of one per $K$ collected transitions and flushed at episode boundaries. Training begins only after 10 episodes have been completed.

\subsubsection{Evaluation Protocol}
\label{sec:app_evaluation}

Each method is evaluated from its final checkpoint unless otherwise specified, under the same latency condition used during training.

\subsection{Baseline Implementations}
\label{sec:app_baselines}

\paragraph{RLPD~\citep{ball2023efficientonlinereinforcementlearning}}
We follow the SERL~\citep{luo2025serlsoftwaresuitesampleefficient} setup for RLPD~\citep{ball2023efficientonlinereinforcementlearning}, using the same $\pi_{0.5}$-style observation pipeline as our real-robot stack. No VLA is used in the policy. The policy is a tanh-Gaussian distribution over a single 7-dimensional action and is queried at every control step. It runs with zero inference delay, so its Bellman backup uses $\gamma$ per environment step rather than $\gamma^C$.

We use hidden dimensions $(256,256,256)$, Adam with a learning rate of $3\times10^{-4}$, $\gamma=0.99$ ($0.997$ for Kick), minibatch size 256, UTD ratio 4, initial temperature 0.1, target entropy $-7/2=-3.5$, and no entropy term in the Bellman backup. The critic uses the same REDQ-style ensemble as \methodname, with 10 networks and two subsampled for each target estimate, together with LayerNorm, a 512-dimensional image latent, and a 64-dimensional state latent. We use the same augmentation procedure, but a smaller ResNetV2 visual backbone with stage depths $(1,1,1,1)$.

Demonstrations constitute 50\% of every critic batch. Training runs in an asynchronous learner thread that is not rate-limited by the environment. We report the measured number of optimizer steps per environment step for each run rather than assuming a fixed ratio.

\paragraph{DSRL~\citep{wagenmaker2025steering}, DSRL w/ RTC}
DSRL uses the frozen $\pi_{0.5}$ policy loaded from the same prefix-conditioned supervised checkpoint used to initialize \methodname. SAC operates in noise space: the actor outputs a single 32-dimensional noise vector corresponding to the padded model action dimension, squashed as $M\tanh(u)$ with $M=1.0$ and tiled across the 16-step action horizon.

The critic is parameterized as $Q(\mathrm{enc}(s),\epsilon)$ and uses the same ResNetV2 encoder as \methodname, with stage depths $(3,4,6,3)$, width 64, input resolution $224\times224$, a 512-dimensional image latent, a 64-dimensional state latent, hidden dimensions $(256,256,256)$, LayerNorm, and a REDQ ensemble of 10 networks with two subsampled for the target. State information is included in the critic, and the same image augmentation is applied.

The actor learning rate is $10^{-4}$, while the critic and temperature use $3\times10^{-4}$. We use $\gamma=0.99$ ($0.997$ for Kick), $\tau_Q=5\times10^{-3}$, minibatch size 64, UTD ratio 20, initial temperature 0.01, target entropy 0, and no entropy term in the Bellman backup.

DSRL executes $C=8$ steps per replan. Under nonzero delay, it uses the same real-time-chunking prefix inpainting as \methodname, with SAC noise applied only to the postfix. The Bellman backup requires no modification because DSRL operates entirely in noise space. DSRL is trained purely online without demonstration data because demonstrations do not contain the corresponding noise labels. For DSRL w/ RTC, we use the same delay as \methodname and same optimization parameters as DSRL.

\paragraph{EXPO-FT~\citep{dong2026expoftsampleefficientreinforcementlearning}, EXPO-FT w/ RTC}
EXPO-FT uses the same learner architecture, network sizes, candidate counts, filter, optimizer settings, and prior-data configuration as \methodname. The only difference is that EXPO-FT is delay-unaware: it is trained and evaluated with $d=0$ while incurring $100$\,ms of real inference latency. This setting isolates the contribution of delay-aware chunking from the underlying EXPO optimization. For EXPO-FT with RTC, we use the same delay and optimization parameters as \methodname.
}

\end{document}